\documentclass[letterpaper]{article} 
\usepackage{aaai23}  
\usepackage{times}  
\usepackage{helvet}  
\usepackage{courier}  
\usepackage[hyphens]{url}  
\usepackage{graphicx} 
\usepackage{natbib}  
\usepackage{caption} 
\usepackage{algorithm}
\usepackage{algorithmic}

\usepackage{newfloat}
\usepackage{listings}
\DeclareCaptionStyle{ruled}{labelfont=normalfont,labelsep=colon,strut=off} 
\floatstyle{ruled}
\newfloat{listing}{tb}{lst}{}
\floatname{listing}{Listing}
\usepackage{amssymb, amsmath, mathtools}
\usepackage{amsthm,amssymb}
\usepackage{adjustbox} 
\usepackage{comment}
\usepackage{enumitem} 
\usepackage{capt-of}

\usepackage{booktabs}

\DeclareMathOperator*{\argmax}{arg\,max}
\DeclareMathOperator*{\argmin}{arg\,min}
\DeclareMathOperator{\E}{\mathop{\mathbb{E}}}
\DeclareMathOperator{\R}{\mathop{\mathbb{R}}}

\title{Learning Pessimism for Reinforcement Learning}
\author {
    Edoardo Cetin,\textsuperscript{\rm 1}
    Oya Celiktutan \textsuperscript{\rm 1}
}
\affiliations{
    \textsuperscript{\rm 1} King's Colloge London\\
    edoardo.cetin@kcl.ac.uk, oya.celiktutan@kcl.ac.uk
}

\begin{document}
\frenchspacing
\maketitle

\begin{abstract}
Off-policy deep reinforcement learning algorithms commonly compensate for overestimation bias during temporal-difference learning by utilizing pessimistic estimates of the expected target returns. In this work, we propose \textit{Generalized Pessimism Learning (GPL)}, a strategy employing a novel \textit{learnable} penalty to enact such pessimism. In particular, we propose to learn this penalty alongside the critic with \textit{dual TD-learning}, a new procedure to estimate and minimize the magnitude of the target returns bias with trivial computational cost. \textit{GPL} enables us to accurately counteract overestimation bias \textit{throughout} training without incurring the downsides of overly pessimistic targets. By integrating \textit{GPL} with popular off-policy algorithms, we achieve \textit{state-of-the-art} results in both competitive proprioceptive and pixel-based benchmarks. 
\end{abstract}

\section{Introduction}
\label{introduction}

Sample efficiency and generality are two directions in which reinforcement learning (RL) algorithms are still lacking, yet, they are crucial for tackling complex real-world problems \citep{rl-real-world-design}. Consequently, many RL milestones have been achieved through simulating large amounts of experience and task-specific parameter tuning \citep{dqn, alphazero}. Recent off-policy model-free \citep{redq} and model-based algorithms \citep{mbpo} advanced RL's sample-efficiency on several benchmark tasks. We attribute such improvements to two main linked advances: more expressive models to capture uncertainty and better strategies to counteract detrimental biases from the learning process. These advances yielded the stabilization to adopt more aggressive optimization procedures, with particular benefits in lower data regimes.

Modern policy gradient algorithms learn behavior by optimizing the expected performance as predicted by the \textit{critic}, a trained parametric model of the agent's performance in the environment. Within this process, overestimation bias naturally arises from the maximization performed over the critic's performance predictions, and consequently, also over the critic's possible errors. In the context of off-policy RL, the critic is trained to predict the agent's future returns via temporal difference (TD-) learning. A common strategy to counteract overestimation is to parameterize this model with multiple, independently-initialized networks and optimize the agent's behavior over the minimum value of the relative outputs \citep{td3}. Empirically, this strategy consistently yields pessimistic target performance measures, avoiding overestimation bias propagating through the TD-learning target bootstraps. However, this approach directly links the critic's parameterization to bias counteraction and appears to suffer from suboptimal exploration due to \textit{underestimation} bias \citep{optadv-OAC}.

Based on these observations, we propose \textit{\textbf{Generalized Pessimism Learning (GPL)}}, a new strategy that \textit{learns} to counteract biases by optimizing a new dual objective. GPL makes use of an explicit penalty to correct the critic's target predictions. We design this penalty as a weighted function of epistemic uncertainty, computed as the expected Wasserstein distance between the return distributions predicted by the critic. We learn the penalty's weight with \textit{dual TD-learning}, a new procedure to estimate and counteract any arising bias in the critic's predictions with dual gradient descent. GPL is the first method to freely learn unbiased performance targets throughout training.

We extend GPL by introducing \textit{pessimism annealing}, a new procedure motivated by the principle of \textit{optimism in the face of uncertainty} \citep{optadv-justOrig}.
This procedure leads the agent to adopt risk-seeking behavior, by utilizing a purposely biased estimate of the performance in the initial training stages. This allows it to trade expected immediate performance for improved directed exploration, incentivizing the visitation of states with high uncertainty from which it would gain more information.

We incorporate GPL with the \textit{Soft Actor-Critic (SAC)} \citep{sac, sac-alg} and \textit{Data-regularized Q (DrQ)} algorithms \citep{drq, drqv2}, yielding \textit{GPL-SAC} and \textit{GPL-DrQ}. On challenging Mujoco tasks from OpenAI Gym \citep{mujoco, gym}, GPL-SAC outperforms both model-based \citep{mbpo} and model-free \citep{redq} state-of-the-art algorithms, while being more computationally efficient. For instance, in the Humanoid environment GPL-SAC recovers a score of 5000 in less than 100K steps, more than nine times faster than regular SAC. Additionally, on pixel-based environments from the DeepMind Control Suite \citep{dmc}, GPL-DrQ provides significant performance improvements from the recent state-of-the-art \textit{DrQv2} algorithm. We validate the statistical significance of our improvements using \textit{Rliable} \citep{precipice-rliable}, further highlighting the effectiveness and applicability of GPL. We share our code to facilitate future extensions.

In summary, we make three main contributions toward improving off-policy reinforcement learning:
\begin{itemize}
    \item We propose Generalized Pessimism Learning, using the first dual optimization method to estimate and precisely counteract overestimation bias 
    throughout training.
    \item To improve exploration, we extend GPL with pessimism annealing, a strategy that exploits epistemic uncertainty to actively seek for more informative states. 
    \item We integrate our method with SAC and DrQ, yielding new state-of-the-art results with trivial computational overheads on both proprioceptive and pixel tasks.
\end{itemize}

\section{Related Work}
\label{related work}

Modern model-free off-policy algorithms utilize different strategies to counteract overestimation bias arising in the critic's TD-targets \citep{over-sem-first, over-sem-0, over-sem-1}. Many approaches combine the predictions of multiple function approximators to estimate the expected returns, for instance, by independently selecting the bootstrap action \citep{double-q-sem}. In discrete control, such a technique appears to mitigate the bias of the seminal \textit{DQN} algorithm \citep{dqn}, consistently improving performance \citep{double-q, rainbow}. In continuous control, similar strategies successfully stabilize algorithms based on the policy gradient theorem \citep{dpg}. Most notably, \citet{td3} proposed to compute the critic's TD-targets by taking the minimum over the outputs of two different action-value models. This minimization strategy has become ubiquitous, being employed in many popular subsequent works \citep{sac-alg}.
For a better trade-off between optimism and pessimism, \citet{weighted-double-q} proposed using a weighted combination of the original and minimized targets. Instead, \citet{dropquant} proposed to parameterize a distributional critic and drop a fixed fraction of the predicted quantiles to compute the targets. Alternative recently proposed strategies for bias-counteraction also entail combining the different action-value predictions with a Softmax function \citep{cr-aaai-softmax-deep-double-q} and computing action-value targets with convex combinations of predictions obtained from multiple actors \citep{cr-aaai-double-actors}. Similarly to our approach, several works considered explicit penalties based on heuristic measures of epistemic uncertainty \citep{bias-corr-q-learn, optadv-OAC}. Recently, \citet{discor} proposed to complement these strategies by further reducing bias propagation through actively weighing the TD-loss of different experience samples. All these works try to \textit{hand-engineer} a fixed penalization to counteract the critic's bias. In contrast, we show that any fixed penalty would be inherently suboptimal (Section~\ref{subsec:dual_td}) and propose a novel strategy to precisely adapt the level of penalization throughout training.

Within model-based RL, 
recent works have achieved remarkable sample efficiency by learning large ensembles of dynamic models for better predictions \citep{handful-of-trials, POPLIN, mbpo}. In the model-free framework, prior works used large critic ensembles for more diverse scopes. \citet{average-dqn} proposed to build an ensemble using several past versions of the value network to reduce the magnitude of the TD-target's bias.  Moreover, \citet{maxmin-q-learning} showed that different sampling procedures for the critic's ensemble predictions can regulate underestimation bias. Their work was extended to the continuous setting by \citet{redq}, which showed that large ensembles combined with a high update-to-data ratio can outperform the sample efficiency of contemporary model-based methods. Ensembling has also been used to achieve better exploration following the optimism in the face of uncertainty principle 
in both discrete \citep{boot-dqn-ucb} and continuous settings \citep{optadv-OAC}. In addition to these advantages, we show that GPL can further exploit large ensembles to better estimate and learn to counteract bias.

In the same spirit as this work, multiple prior methods attempted to learn the components and parameters of underlying RL algorithms. Several works have approached this problem by utilizing expensive meta-learning strategies to obtain new learning objectives based on the multi-task performance from low-computation environments \citep{autoRL0meta, autoRL1meta, autoRL-1meta}. 
More related to our method, \textit{Tactical Optimism and Pessimism} \citep{top} introduced the concept of adapting a bias penalty online. Together with similar later work \citep{autoTQC}, they proposed step-wise updates to the bias correction parameters based on the performance of recent trajectories. Instead, GPL proposes a new method to precisely estimate bias and reduce its magnitude via dual gradient descent. We provide a direct empirical comparison and further motivate the advantages of our approach in App.~\ref{appsub:alt_opt_beta}.%
\section{Preliminaries}

\label{sec:prel}

In RL, we aim to autonomously recover optimal agent behavior for performing a particular task. Formally, we describe this problem setting as a Markov Decision Process (MDP), defined as the tuple $(S, A, P, p_0, r, \gamma)$.
At each time-step of interaction, the agent observes some state in the state space, $s\in S$, and performs some action in the action space, $a \in A$. The transition dynamics function $P: S \times A \times S \rightarrow \mathbb{R}$ and the initial state distribution $p_0: S\rightarrow \mathbb{R}$ describe the evolution of the environment as a consequence of the agent's behavior. The reward function $r: S \times A \rightarrow \mathbb{R}$ quantifies the effectiveness of each performed action, while the discount factor $\gamma\in [0, 1]$ represents the agent's preference for earlier rewards. A policy $\pi: S \times A \rightarrow \mathbb{R}$ maps each state to a probability distribution over actions and represents the agent's behavior. An episode of interactions between the agent and the environment yields a trajectory $\tau$ containing the transitions experienced, $\tau = (s_0, a_0, r_0, s_1, a_1, r_1, ...)$. The RL objective is then to find an optimal policy $\pi^*$ that maximizes the expected sum of discounted future rewards:
\begin{equation}
\small
\label{pi_obj}
\pi^*=\argmax_\pi\E_{p_\pi(\tau)}\left[ \sum^{\infty}_{t = 0}\gamma^t r (s_t, a_t) \right],
\end{equation}
where $p_\pi(\tau)$ represents the distribution of trajectories stemming from the agent's interaction with the environment. Off-policy RL algorithms commonly utilize a critic model to evaluate the effectiveness of the agent's behavior. A straightforward choice for the critic is to represent the policy's action-value function $Q^\pi: S \times A \rightarrow \mathbb{R}$. This function quantifies the expected sum of discounted future rewards after executing some particular action from a given state:
\begin{equation}
\small
\label{q_fn}
Q^\pi(s, a) = \E_{p_\pi(\tau|s_0{=}s, a_0{=}a)}\left[ \sum^{\infty}_{t = 0}\gamma^t r (s_t, a_t)\right].
\end{equation}
Most RL algorithms consider learning parameterized models for both the policy, $\pi_\theta$, and the corresponding action-value function, $Q^\pi_\phi$. In particular, after storing experience transitions $(s, a, s', r)$ in a replay data buffer $D$, we learn $Q^\pi_\phi$ 
by iteratively minimizing a squared TD-loss of the form:
\begin{equation}\label{q_fn_obj}
\small
\begin{split}
    J_Q(\phi)=\E_{s, a, s', r \sim D}\left[(Q^\pi_\phi(s, a) - y)^2\right],\\
    y= r + \gamma\mathbb{E}_{a\sim \pi(s')}\left[\hat{Q}_{\phi'}^\pi(s', a)\right].
\end{split}
\end{equation}
Here, the TD-targets $y$ are obtained by computing a 1-step bootstrap with a \textit{target action-value estimator} $\hat{Q}_{\phi'}^\pi$. Usually, $\hat{Q}_{\phi'}^\pi$ is a regularized function of action-value predictions from a target critic model using delayed parameters $\phi'$. 
Following the policy gradient theorem \citep{pg-thm, dpg}, we can then improve our policy by maximizing the expected returns as predicted by the critic. This corresponds to minimizing the negation of the critic's expected target action-value estimates:
\begin{equation}
\small
\label{pi_pg_obj}
J_\pi(\theta)=-\E_{s\sim D, a\sim \pi_\theta(s)}\left[\hat{Q}_\phi^\pi(s,a)\right].
\end{equation}

\section{Addressing Overestimation Bias}
\label{sec: analysis}

\begin{figure*}[t]
  \centering
  \includegraphics[width=0.85\textwidth]{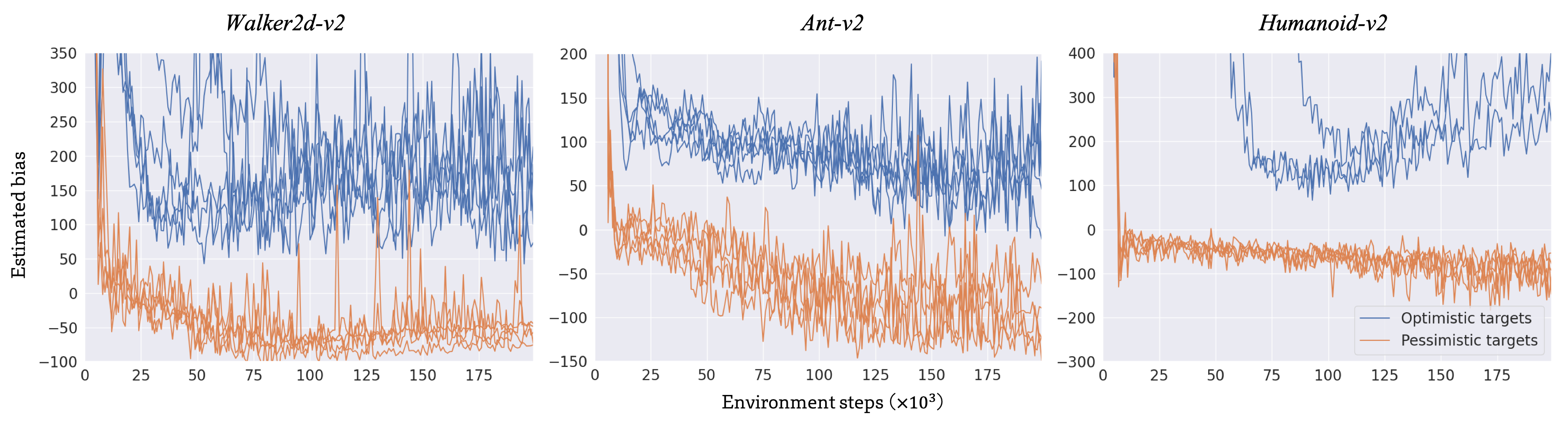}
  \caption{Recorded estimated bias for ten runs of two simple extensions of the \textit{SAC} algorithm.} 
  \label{figure:mujoco_bias}
\end{figure*}

\subsection{Bias in Q-Learning}
\label{sec:q-bias}
In off-policy RL, several works have identified an accumulation of overestimation bias in the action-value estimates as a consequence of TD-learning \citep{over-sem-first, over-sem-1}. Formally, we quantify the target action-value bias $B(s, a, s')$ as the difference between the actual and estimated TD-targets of a transition \citep{redq}:
\begin{equation}
\label{bias_obj}
\begin{split}
\small 
    B(s, a, s')
    =\gamma\mathbb{E}_{a'\sim \pi(s')}\left[\hat{Q}_{\phi'}^\pi(s', a') - Q^\pi(s', a')\right].
\end{split}
\end{equation}
Positive bias arises when the target action-values are obtained directly from the outputs of a parameterized action-value function, i.e., $\hat{Q}_{\phi'}^\pi=Q^\pi_{\phi'}$ \citep{td3}. The reason for this phenomenon is that the policy is trained to locally maximize the action-value estimates from Eqn.~\ref{pi_pg_obj}. Hence, its actions will exploit potential model errors to obtain higher scores, implying that $\E_{s, a\sim\pi(s)}[Q^\pi_{\phi'}(s, a)]>\E_{s, a\sim\pi(s)}[Q^\pi(s, a)]$. Instabilities then arise as the errors can quickly propagate through the bootstrap operation, inherently causing the phenomenon of \textit{positive bias accumulation}. To counteract this phenomenon, \citet{td3} proposed \textit{clipped double Q-learning}. This technique consists in learning two separate action-value functions and computing the target action-values by taking the minimum over their outputs:
\begin{equation}\label{eq:clippeddoubleQ}
\small
    \hat{Q}_{\phi'_{min}}^\pi(s, a)=\min\left(Q_{\phi_1'}^\pi(s, a), Q_{\phi_2'}^\pi(s, a)\right).
\end{equation}
The role of the minimization is to consistently produce overly pessimistic estimates of the target action-values, preventing positive bias accumulation. This approach is an empirically effective strategy for different benchmark tasks and has become standard practice.

\subsection{The Uncertainty Regularizer}

\label{sec:gen_unc_reg}
 In this work, we take a more general approach for computing the target action-values. Particularly, we use a parameterized function, the \textit{uncertainty regularizer} $p_\beta(s, a, \phi, \theta)$, for trying to approximate the bias in the critic's action-value predictions for on-policy actions. 
Thus, we specify an action-value estimator that penalizes the action-value estimates via the uncertainty regularizer:
\begin{equation}\label{q_pen_ver}
\small
\begin{split}
&\hat{Q}_{\phi'}^\pi(s, a|\beta)=Q_{\phi'}^\pi(s, a) - p_\beta(s, a, \phi', \theta),\\
\textit{wh}&\textit{ere} \quad p_\beta(s, a, \phi', \theta) \approx Q_{\phi'}^\pi(s, a) - Q^\pi(s, a).
\end{split}
\end{equation}
\normalsize
A consequence of this formulation is that as long as $p_\beta$ is unbiased \textit{for on-policy actions}, so will the action-value estimator $\hat{Q}_{\phi'}^\pi$. 
Therefore, this would ensure that the expected target action-value bias is zero, preventing the positive bias accumulation phenomenon without requiring overly pessimistic action-value estimates. Based on these observations, 
we now specify a new method that learns an unbiased $p_\beta$ and continuously adapts it to reflect changes in the critic and policy.

\section{Generalized Pessimism Learning}
\label{sec:method}

Generalized Pessimism Learning (GPL) entails learning a particular uncertainty regularizer $p_\beta$, which we precisely specify in Eqn.~\ref{wass_pen}. Our approach makes $p_\beta$ adapt to changes in both actor and critic models throughout the RL process, to keep the target action-values unbiased. Hence, GPL allows for preventing positive bias accumulation without overly pessimistic targets. With any fixed penalty, we argue that it would be infeasible to maintain the expected target action-value bias close to zero due to the number of affecting parameters and stochastic factors in different RL experiments.

\subsection{Uncertainty Regularizer Parameterization}

We strive for a parameterization of the uncertainty regularizer that ensures low bias and variance estimation of the target action-values. Similar to prior works \citep{optadv-OAC, top}, GPL uses a linear model of some measure of the epistemic uncertainty in the critic. Epistemic uncertainty represents the uncertainty from the model's learned parameters towards its possible predictions. Hence, when using expressive deep models, the areas of the state and action spaces where the critic's epistemic uncertainty is elevated are the areas in which the agent did not yet observe enough data to reliably predict its returns and, for this reason, the magnitude of the critic's error is expectedly higher. Consequently, if a policy yields behavior with high epistemic uncertainty in the critic, it is likely exploiting positive errors and overestimating its expected returns. As we use the policy to compute the TD-targets, the higher the uncertainty, the higher the expected positive bias. 


We propose measuring epistemic uncertainty with the expected Wasserstein distance between the critic's predicted return distributions $Z^\pi$ \citep{distributional-q-belle}.
In our main experiments we consider the usual non-distributional case where we parameterize the critic with multiple action-value functions, in which case we view each action-value estimate as a Dirac delta function approximation of the return distribution, $Z^\pi_\phi(s, a)=\delta_{Q^\pi_\phi(s, a)}$. Our uncertainty regularizer then consists of linearly scaling the expected Wasserstein distance via a learnable parameter $\beta$:
\begin{equation}\label{wass_pen}
\small
\small
p_\beta(s, a, \phi, \theta)= \beta\times \E_{a, \phi_1, \phi_2}\left[W(Z^\pi_{\phi_1}(s, a), Z^\pi_{\phi_2}(s, a))\right].
\end{equation}
We estimate the expectation in Eqn.~\ref{wass_pen} by learning 
$N\geq2$ independent critic models with parameters $\{\phi_i\}_{i=1}^N$, and averaging the distances between the corresponding predicted return distributions. Notably, the Wasserstein distance has easy-to-compute closed forms for many popular distributions. For Dirac delta functions, it is equivalent to the distance between the corresponding locations, hence, $W(\delta_{Q^\pi_{\phi_1}(s, a)}, \delta_{Q^\pi_{\phi_2}(s, a)}) = |Q^\pi_{\phi_1}(s, a)-Q^\pi_{\phi_2}(s, a)|$.

Our quantification of epistemic uncertainty is an interpretable measure for any distributional critic. Moreover, for some fixed $\beta$, increasing the number of critics decreases the estimation variance but leaves the expected magnitude of the uncertainty regularizer unchanged. This is because the sample mean of the Wasserstein distances is always an unbiased estimate of Eqn.~\ref{wass_pen} for $N\geq 2$. Assuming we can approximately model the distribution of different action-value predictions with a Gaussian, we can show our penalty is proportional to the standard deviation of the distribution of action-value predictions. We can also restate clipped double Q-learning using our uncertainty regularizer with $N=2$ and $\beta=0.5$, allowing us to replicate its penalization effects for $N>2$ by simply fixing $\beta$. In contrast, \citet{optadv-OAC} proposed the sample standard deviation of the action-value predictions to quantify epistemic uncertainty. However, the sample standard deviation does not have a clear generalization to arbitrary distributional critics and its expected magnitude is dependent on the number of models. We provide all formal derivations for the above statements in App.~\ref{app:pen}. 

\begin{figure*}[t]
  \centering
  \includegraphics[width=0.85\textwidth]{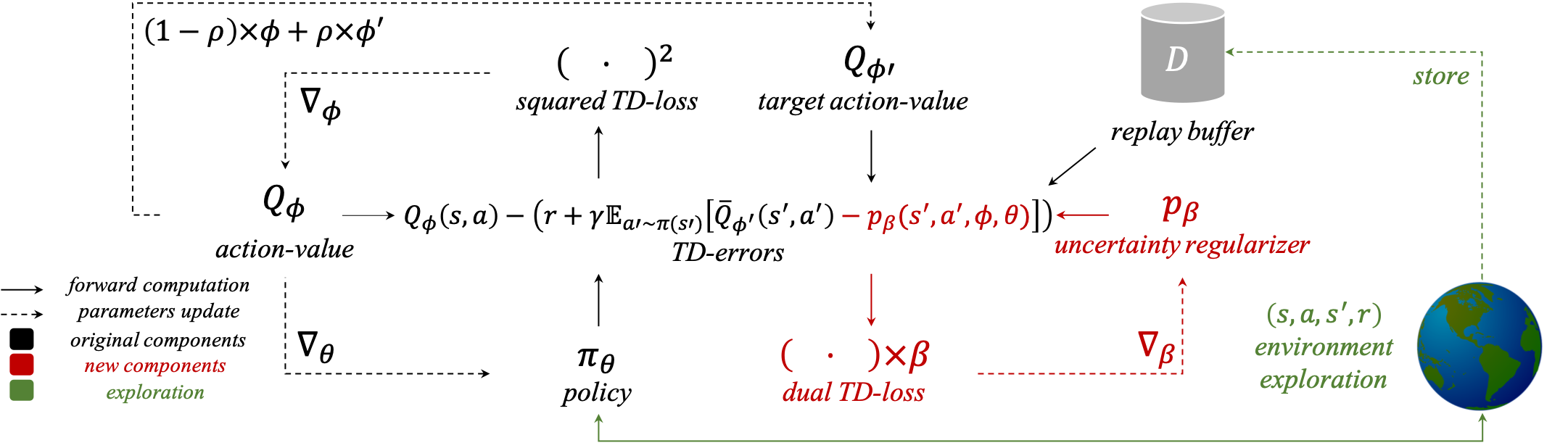}
  \caption{Schematic overview of the training and exploration processes involved in the 
  GPL framework. The TD-errors play a \textit{central role} both for updating the critic and for estimating the current bias to update the uncertainty regularizer.}
  \label{figure:schem-ov}
\end{figure*}

\subsection{Dual TD-Learning}
\label{subsec:dual_td}
We hypothesize that the expected bias present in the action-value targets is highly dependent on several unaccountable factors from the stochasticity in the environment and the learning process. This \textit{extends} recent results showing that the effectiveness of any bias penalty highly varies across tasks (\citet{top}, Fig. 4).
We empirically validate our hypothesis by running multiple experiments with simple extensions of the SAC algorithm in different Gym environments and periodically recording estimates of the action-value bias by comparing the actual and estimated discounted returns (as described in App. \ref{app:subsection:empirical_bias_estimation_desc}).
As shown in Figure~\ref{figure:mujoco_bias}, the bias in the predicted action-values notably varies across environments, agents, training stages, and even across different random seeds. These results validate our thesis that there is no \textit{fixed} penalty able to account for the many sources of stochasticity in the RL process, even for a single task. Hence, this shows the necessity of learning $p_\beta$ alongside the policy and critic to accurately counteract bias.

When using the uncertainty regularizer, we will denote the target bias for a transition as $B(s, a, s'|\beta)$, to highlight its dependency on the current value of $\beta$. Furthermore, note that $B(s, a, s'|\beta)$ would take on a positive or negative value in the cases where $\beta$ yields either insufficient or excessive regularization, respectively. Therefore, to recover unbiased targets, we propose to optimize $\beta$ as a dual variable by enforcing the expected target action-value bias to be zero:
\begin{equation}\label{ext_q_fn_obj}
\small 
\begin{split}
    \argmin_{\beta} -\beta\times \E_{s, a, s'\sim D}\left[B(s, a, s'|\beta)\right].
\end{split}
\end{equation}
To estimate $B(s, a, s'|\beta)$, we use the property that for \textit{arbitrary off-policy} actions the action-value estimates are not directly affected by the positive bias from the policy gradient optimization \citep{td3}.
Consequently, we make the 
assumption that $Q^\pi_\phi$ itself provides \textit{initially} unbiased estimates of the expected returns, i.e., $\E_{s, a\sim D}[Q^\pi_\phi(s, a)]\approx \E_{s, a\sim D}[Q^\pi(s, a)]$. This assumption directly implies that any expected error in the Bellman relationship between $r + \gamma\mathbb{E}_{a'\sim \pi(s')}\left[\hat{Q}_{\phi'}^\pi(s', a'|\beta)\right]$ and $Q^\pi_\phi(s, a)$ is due to bias present in our action-value estimator. Hence, we propose to 
approximate $B(s, a, s'|\beta)$, with the expected difference between the current \textit{on-policy} TD-targets and action-value predictions for \textit{off-policy} actions:
\begin{equation}\label{approx_bias}
\small 
B(s, a, s'|\beta) \approx 
r + \gamma\mathbb{E}_{a'\sim \pi(s')}\left[\hat{Q}_{\phi'}^\pi(s', a'|\beta)\right] - Q^\pi_\phi(s, a).
\end{equation}
In practice, GPL alternates the optimizations of $\beta$ for the current bias, and both actor and critic parameters, with the corresponding updated RL objectives. This is similar to the automatic exploration temperature optimization proposed by \citet{sac-alg}, approximating dual gradient descent \citep{convex}. We can estimate the current bias according to Eqn.~\ref{approx_bias} by simply negating the already-computed errors from the TD-loss, with trivial cost. Thus, we name this procedure \textit{dual TD-learning}. We provide a schematic overview of GPL with dual TD-learning in Fig.~\ref{figure:schem-ov}.


\textbf{Limitations.} When using deep networks and approximate stochastic optimization, we recognize that the unbiasedness assumption of Eqn.~\ref{approx_bias} might not necessarily hold. Therefore, given initially biased action-value targets, some of the bias might propagate to the critic model, influencing the approximation in Eqn.~\ref{approx_bias}. This property of our method makes it hard to provide a formal analysis beyond the tabular setting. However, in practice, we still find that GPL's performance and the optimization dynamics are robust to different levels of initial target bias, and that dual TD-learning appears to always improve over fixed penalties. We show this in App.~\ref{ext:unc_reg}, by analyzing the empirical behavior and performance of GPL with different initial values of $\beta$ and, thus, different bias in the initial targets. An intuition for our results is that the relative difference between the off-policy and on-policy action-value predictions should always push $\beta$ to counteract new bias stemming from model errors in the 
policy gradient action maximization, 
and thus improve over non-adaptive methods which are also affected by initial bias. We further validate dual TD-learning in App.~\ref{ext:alt_opt_beta}, comparing with optimizing $\beta$ by minimizing the squared norm of the bias 
and by using the bandit-based optimization from TOP \citep{top}. We also note that integrating GPL adds non-trivial complexity by introducing an entirely new optimization step which could be unnecessary for low-dimensional and easy-exploration problems. This inevitable limitation could further exacerbate the reproducibility of off-policy RL \citep{reproducibility-rl}.

\subsection{Pessimism Annealing for Directed Exploration}

\label{sec:act-shift}
As described in Section~\ref{sec:prel}, the policy learns to maximize the unbiased action-values predicted by 
action-value estimator $\hat{Q}^\pi_{\phi}$. Motivated by the principle of optimism in the face of uncertainty \citep{optadv-justOrig}, we propose to make use of a new 
\textit{optimistic} policy gradient objective:
\begin{equation}
\small 
\begin{split}
    &J^{opt}_\pi(\theta) = -\E_{s\sim D, a\sim \pi(s)}[\hat{Q}^{\pi_{opt}}_{\phi}(s, a|\beta)],\\
    \textit{where}&\quad \hat{Q}^{\pi_{opt}}_{\phi}(s, a|\beta)=Q_{\phi}^\pi(s, a) - p_{\beta_{opt}}(s, a, \phi, \theta).
\end{split}
\end{equation}
This objective utilizes an \textit{optimistic shifted uncertainty regularizer}, $p_{\beta_{opt}}$, calculated with parameter $\beta_{opt} = \beta - \lambda_{opt}$, for a decaying \textit{optimistic shift value}, $\lambda_{opt} \geq 0$. This new objective trades off the traditional exploitative behavior of the policy with directed exploration. As $\lambda_{opt}$ is large, $\pi$ will be incentivized to perform actions for which the outcome has high epistemic uncertainty. Therefore, the agent will experience transitions that are increasingly informative for the critic but expectedly sub-optimal. Hence, we name the process of decaying $\lambda_{opt}$ \textit{pessimism annealing}, enabling to achieve improved exploration early on in training without biasing the policy's final objective.

\begin{figure*}[t]
  \centering
  \includegraphics[width=0.98\textwidth]{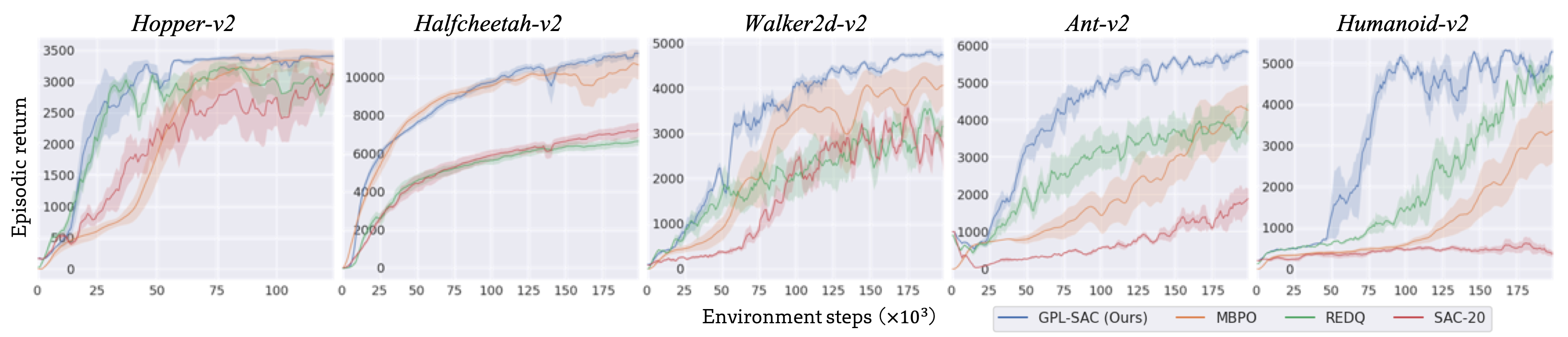}
  \caption{Performance curves for the considered complex Mujoco environments from OpenAI Gym,  over five random seeds.}
  \label{figure:mujoco_res}
\end{figure*}

\section{Experiments}
\label{sec:experiments}

\label{sec:exp_prop}

To evaluate the effectiveness of GPL, we integrate it with two popular off-policy RL algorithms. GPL itself introduces trivial computational and memory costs as it optimizes a single additional weight, re-utilizing the errors in the TD-loss to estimate the bias. Moreover, we implement the critic's ensemble as a single neural network, using linear non-fully-connected layers evenly splitting the nodes and dropping the weight connections between the splits. Practically, when evaluated under the same hardware, this results in our algorithm running more than 2.4 times faster than the implementation from \citet{redq} while having a similar algorithmic complexity (see App.~\ref{ext:comp_scaling}). 

 We show that GPL significantly improves the performance and robustness of off-policy RL, concretely surpassing prior algorithms and setting new state-of-the-art results. In our evaluation, we repeat each experiment with five random seeds and record both mean and standard deviation over the episodic returns. Moreover, we validate statistical significance using tools from \textit{Rliable}~\citep{precipice-rliable}. We report all details of our experimental settings and utilized hyper-parameters in App.~\ref{app:params}. Furthermore, we provide comprehensive extended results analyzing the impact of all relevant design choices, testing several alternative implementations, and reporting all training times in App.~\ref{app:ext_emp_ana}.

\subsection{Continuous control from proprioception}


\textbf{GPL-SAC.} First, we integrate GPL as a plug-in addition to \textit{Soft Actor-Critic (SAC)} \citep{sac, sac-alg}, a popular model-free off-policy algorithms that uses a weighted entropy term in its objective to incentivize exploration. Specifically, we only substitute SAC's clipped double Q-learning with our uncertainty regularizer, initialized with $\beta=0.5$. Inline with the other considered state-of-the-art baselines \citep{redq,mbpo}, we use an increased ensemble size and update-to-data (UTD) ratio for the critic. We found both these choices necessary for sample-efficient learning in the evaluated experience regimes (App~\ref{appsub:ens}-\ref{appsub:utd}). We denote the resulting algorithm GPL-SAC and summarize this integration in Algorithm 1 (App. B). We would like to note that all other practices, unrelated to counteracting overestimation bias (such as learning the entropy bonus) were already present in SAC and are utilized by all baselines.

\textbf{Baselines.} We compare \textbf{\emph{GPL-SAC}} with prior state-of-the-art model-free and model-based algorithms with similar or greater computational complexity, employing high UTD ratios: \textbf{\emph{REDQ}} \citep{redq}, state-of-the-art model-free algorithm on OpenAI Gym. This algorithm learns multiple action-value functions and utilizes clipped double Q-learning over a sampled pair of outputs to compute the critic's targets. \textbf{\emph{MBPO}} \citep{mbpo}, state-of-the-art, model-based algorithm on OpenAI Gym. This algorithm learns a large ensemble of world models with \textit{Dyna}-style \citep{dyna} optimization to train the policy. \textbf{\emph{SAC-20}}, simple SAC extension where with an increased UTD ratio of 20.

\textbf{Results.} We evaluate GPL-SAC compared to the described baselines on five of the more challenging Mujoco environments from OpenAI Gym \citep{gym}, involving complex locomotion problems from proprioception. 
We 
collect the returns over five evaluation episodes every 1000 environment steps. In Figure~\ref{figure:mujoco_res}, we 
show the different performance curves. 
GPL-SAC is consistently the best performing algorithm on all environments, setting new state-of-the-art results for this benchmark at the time of writing.
Moreover, the performance gap is greater for tasks with larger state and action spaces. 
We motivate this by noting that increased task-complexity appears to correlate with an increased stochasticity affecting target bias (Fig.~\ref{figure:mujoco_bias}), increasing the necessity for an adaptive counteraction strategy. Furthermore, as all baselines use fixed strategies to deal with overestimation bias, they also require overly pessimistic estimates of the returns to avoid instabilities. Hence, the resulting policies are likely overly conservative, hindering exploration and efficiency, with larger effects on higher-dimensional tasks. For instance, on \textit{Humanoid}, GPL-SAC remarkably surpasses a score of 5000 after only 100K steps, more than $9\times$ faster than SAC and $2.5\times$ faster than REDQ.

\begin{table}[t]
\caption{Results summary for the pixel observations experiments on the DeepMind Control Suite. We provide full per-environment results and performance curves in App.~\ref{app:res_vis_dmc}} \label{tab:gpldrq}

\vskip 0.1in
\tabcolsep=0.07cm
\begin{center}
\adjustbox{max width=0.95\columnwidth}{
\begin{tabular}{@{}lccccc@{}}
\toprule
\textbf{Metric\textbackslash{}Algorithm} & GPL-DrQ+Anneal                       & GPL-DrQ & DrQv2  & CURL   & SAC   \\ \midrule
\textbf{Evaluation milestone}            & \multicolumn{5}{c}{1.5M frames}                                         \\ \cmidrule(l){2-6} 
\textbf{Average score}                   & \textbf{640.20}                     & 620.24  & 544.67 & 302.74 & 50.34 \\
\textbf{Top score count}                 & \textbf{11/12}                      & 8/12    & 3/12   & 0/12   & 0/12  \\ \midrule
\textbf{Evaluation milestone}            & \multicolumn{5}{c}{3.0M frames}                                         \\ \cmidrule(l){2-6} 
\textbf{Average score}                   & \textbf{744.09} & 720.29  & 670.95 & 318.38 & 59.38 \\
\textbf{Top score count}                 & \textbf{10/12}  & 7/12    & 4/12   & 0/12   & 0/12  \\ \bottomrule
\end{tabular}
}

\end{center}
\end{table}

\subsection{Continuous Control from Pixels}

\begin{figure*}[t]
  \centering
  \includegraphics[width=0.99\textwidth]{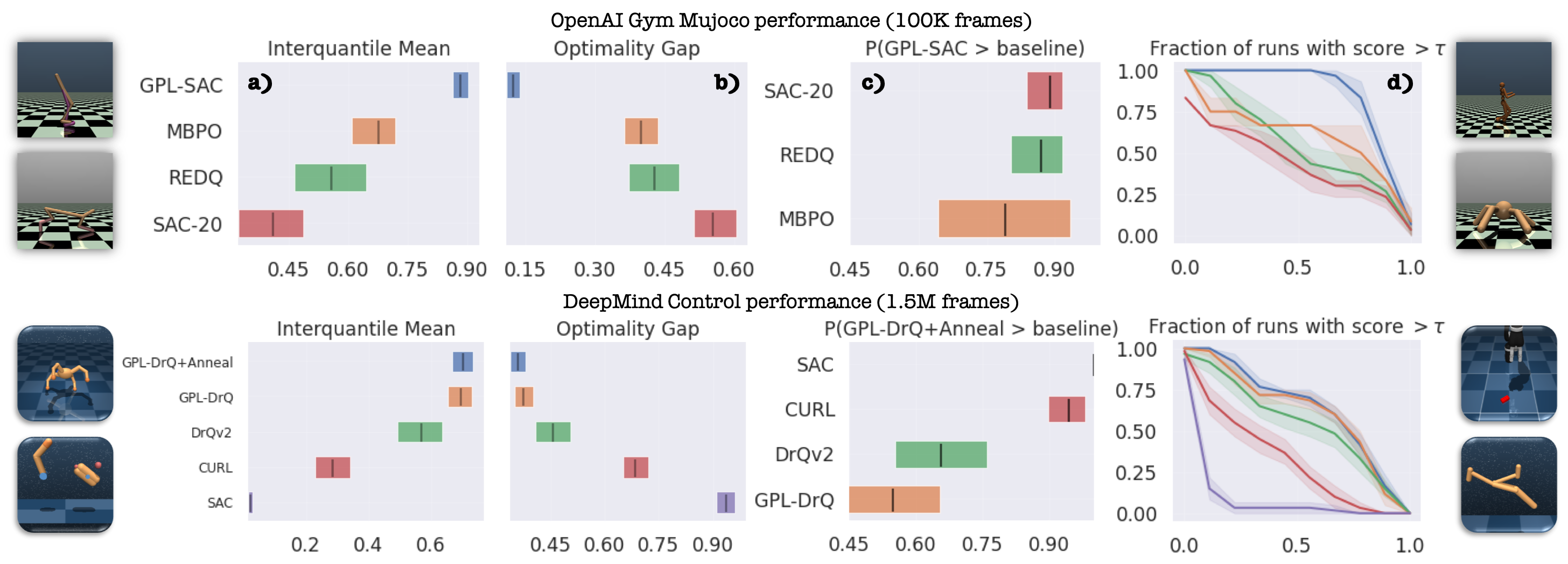}
  \caption{Aggregate performance metrics using the evaluation protocol proposed by \textit{Rliable}.
  We report a) \textit{interquantile mean} 
  b) \textit{optimality gap} 
  c) \textit{probability of improvement}, and d) full \textit{performance profiles}. Ranges/shaded regions correspond to 95\% CIs.} 
  \label{figure:rliable}
\end{figure*}


\textbf{GPL-DrQ.} 
We also incorporate GPL  to a recent version of \textit{Data-regularized Q} (DrQv2) \citep{drqv2}, an off-policy, model-free algorithm achieving state-of-the-art performance for pixel-based control problems. 
DrQv2 combines image augmentation from DrQ \citep{drq} with several advances such as n-step returns \citep{sutton-reinforcement-n-step} and scheduled exploration noise \citep{scheduled-exploration-noise}. Again, we only substitute DrQv2's clipped double Q-learning with our uncertainty regularizer. To bolster exploration, 
we also integrate pessimism annealing from Section~\ref{sec:act-shift}, with $\lambda_{opt}$ linearly decayed from 0.5 to 0.0 together with the exploration noise in DrQv2. We leave the rest of the hyper-parameters and models unaltered to evaluate the generality of applying GPL. We name the resulting algorithms GPL-DrQ and \textit{GPL-DrQ+Anneal}. 

\textbf{Baselines.} We compare our \textbf{\emph{GPL-DrQ}} and \textbf{\emph{GPL-DrQ+Anneal}} with model-free baselines for continuous control from pixels: The aforementioned state-of-the-art \textbf{\emph{DrQv2}}. \textbf{\emph{CURL}} \citep{curl}, recent algorithm combining off-policy learning with a contrastive objective for representation learning. \textbf{\emph{SAC}}, simple integration of SAC with a convolutional encoder.


\textbf{Results.} We evaluate GPL-DrQ and GPL-DrQ+Anneal on the environments from the DeepMind Control Suite \citep{dmc} modified to yield pixel observations. We use the medium benchmark evaluation as described by \citet{drqv2}, consisting of 12 complex tasks involving control problems with hard exploration and sparse rewards. In Table~\ref{tab:gpldrq}, we report the mean returns obtained after experiencing 3M and 1.5M environment frames together with the number of tasks where each algorithm achieves \textit{top scores} within half a standard deviation from the highest recorded return. For each run, we average the returns from 100 evaluation episodes collected in the 100K steps preceding each of these milestones. We provide the full per-environment results and performance curves in App.~\ref{app:res_vis_dmc}. Both GPL-DrQ and GPL-DrQ+Anneal significantly improve the performance of DrQv2 and all other baseline algorithms in the great majority of the tasks (7-11 out of 12). DrQv2 yields inconsistent returns on some tasks, likely due to a lack of exploration from its overly pessimistic critic. 
GPL generally appears to resolve this issue, while pessimism annealing further aids precisely in the tasks where under-exploration is more frequent. Overall, these results show both the generality and effectiveness of GPL for improving the current state-of-the-art through simple integrations, providing a novel framework to better capture and exploit bias.

\subsection{Statistical Significance}

To validate the statistical significance of the performance gains of GPL over the considered state-of-the-art algorithms, we follow the evaluation protocol proposed by \citet{precipice-rliable}. For both Mujoco and DeepMind Control (DMC) benchmarks, we calculate various informative aggregate statistical measures of each algorithm halfway through training, normalizing each task's score within $[0, 1]$ (see App.~\ref{app:params} for details). Error bars/shaded regions correspond to the 95\% stratified bootstrap confidence intervals (CIs) for each algorithm's performance \citep{bootstrap-cis}. As reported in Figure~\ref{figure:rliable}, in both benchmarks GPL achieves considerably \textit{higher interquantile  mean} (a) and \textit{lower optimality gap} (b) than any of the baselines, with non-overlapping CIs. Furthermore, we analyze \textit{probability of improvement} (c) from the Mann-Whitney U statistic \citep{mannwhitneyUstat}, which reveals that GPL's improvements are \textit{statistically meaningful}  as per the Neyman-Pearson statistical testing criterion \citep{stat-significance-neyman-pearson}. Lastly, we calculate the different \textit{performance profiles} (d) \citep{perf-profile}, which show that GPL-based algorithms \textit{stochastically dominate} all considered state-of-the-art baselines \citep{stoch-dom}. We believe these results convincingly validate 
the effectiveness and future potential of GPL.

\section{Discussion and Future Work}

\label{conclusion}


We proposed Generalized Pessimism Learning, a strategy that adaptively \textit{learns} a penalty to recover an unbiased performance objective for off-policy RL. Unlike traditional methods, GPL achieves training stability without necessitating overly pessimistic estimates of the target returns, thus, improving convergence and exploration. We show that integrating GPL with modern algorithms yields state-of-the-art results for both proprioceptive and pixel-based control tasks. Moreover, GPL's penalty has a natural generalization to different distributional critics and variational representations of the weights posterior. Hence, our method has the potential to facilitate research in off-policy reinforcement learning, going beyond action-value functions and model ensembles. Future extensions of our framework could also have significant implications for offline RL, a problem setting particularly sensitive to regulating overestimation bias. 

\bibliography{main}

\appendix

\newpage

\section*{Appendix}

\section{Penalties for Bias Counteraction}
\label{app:pen}

\textit{Generalized Pessimism Learning} makes use of the uncertainty regularizer penalty to counteract biases arising from the off-policy RL optimization process. Here, we further analyze some of the properties and relationships of our penalty with alternatives from prior works.  We will consider the common case where the critic is parameterized as an ensemble of $N\geq 2$ action-value functions $\{Q^\pi_{\phi_i}\}_{i=1}^N$. To simplify our analysis, we assume that the different action-value predictions for a given input are independently drawn from a Gaussian distribution with some variance $\sigma^2_{Q(s,a)}$. We motivate this latter assumption by the \textit{Central Limit Theorem}, given the many sources of stochasticity affecting the training process of each action-value model in the ensemble. To simplify our notation, we will drop input dependencies when doing so will not compromise clarity, e.g., $\sigma^2_{Q(s,a)}\Rightarrow \sigma^2_{Q}$.

\subsection{Expected Penalization of the Uncertainty Regularizer}

The uncertainty regularizer penalty $p_\beta(s, a, \phi, \theta)$ is parameterized as a weighted model of epistemic uncertainty.  Particularly, we propose to measure epistemic uncertainty with the expected Wasserstein distance between the return distributions predicted by the critic. This is formally described in Eqn.~\ref{wass_pen} from Section~\ref{sec:method}. In the common case where we parameterize the critic with an ensemble of action-value functions, we treat each as a Dirac delta function approximation of the return distribution. Consequently, we can estimate the expected Wasserstein distance by averaging over all the $N^2 - N$ absolute differences between different Dirac locations predicted by the action-value functions:
\begin{equation}
\small
\label{eq:wass_approx}
    \begin{split}
    p_\beta(s, a, \phi, \theta) &= \beta\times \E_{a\sim\pi_\theta(s), \phi_1, \phi_2}\left[W(Z^\pi_{\phi_1}(s, a), Z^\pi_{\phi_2}(s, a))\right] \\
        &=\beta\times \E_{a\sim\pi_\theta(s), \phi_1, \phi_2}\left[|Q^\pi_{\phi_1}(s, a)-Q^\pi_{\phi_2}(s, a)|\right] \\
        &\approx \frac{\beta}{N^2-N}\sum_{i=1}^N \sum_{j\neq i}^N |Q_{\phi_i}^{\pi}(s, a) - Q_{\phi_j}^{\pi}(s, a)|.
    \end{split}
\end{equation}
From the independence assumption of the ensemble models, the differences in the action-value predictions will also follow a Gaussian distribution:
\begin{equation}\small
    \begin{split}
    D_{ij} = Q_{\phi_i}^{\pi}(s, a) - Q_{\phi_j}^{\pi}(s, a) \sim N(0, 2\sigma^2_{Q}).
    \end{split}
\end{equation}

Each absolute difference $W_{ij} = |D_{ij}|$ will then follow a \textit{folded normal} distribution \citep{foldednormal}, with moments given by:
\begin{equation}\small
    \begin{split}
    \E[W]  = \mu_W &= \frac{\sqrt{2}}{\pi}\sigma_D = \frac{2}{\pi}\sigma_Q,\\
    var(W)  = \sigma_W &= \sigma^2_D - \mu_W^2 = \left(2-\frac{4}{\pi^2}\right)\sigma_Q^2.
    \end{split}
\end{equation}
Consequently, the expected penalization of the uncertainty regularizer immediately follows:
\begin{equation}\small\label{eq:exp_uncreg}
    \begin{split}
    \E[p_\beta] & = \beta\mu_W =  \frac{2\beta}{\pi}\sigma_Q.\\
    \end{split}
\end{equation}


\subsection{Expected Penalization of the Population Standard Deviation}

\citet{optadv-OAC} and \citet{top} make use of a weighted model of an alternative epistemic uncertainty measure to define a penalty as $p^s_\beta(s, a, \phi, \theta) = \beta\times s(s, a, \phi, \theta)$. In particular, they make use of the population standard deviation $s(s, a, \phi, \theta)$, treating the critic's ensemble predictions as independently sampled action-values:
\begin{equation}\small
\label{eq:pop_std}
    \begin{split}
    s(s, a, \phi, \theta) &=\sqrt{\frac{1}{N} \sum_{i=1}^N\left(Q_{\phi_i}^{\pi}(s, a) - \mu_{Q_\phi(s, a)}\right)^2},\\
    &\textit{where} \quad \mu_{Q_\phi} = \frac{1}{N}\sum_{i=1}^N Q_{\phi_i}^{\pi}.
    \end{split}
\end{equation}
We can rewrite the population standard deviation measure in terms of the square root of the sum of $N$ squared difference terms $D_i$:
\begin{equation}\small
\begin{split}
     s &=\sqrt{\sum_{i=1}^N D_i^2}, \quad \textit{where}\\ D_i &= \frac{1}{\sqrt{N}}\left(Q_{\phi_i}^{\pi} - \mu_{Q_\phi}\right).   
\end{split}
\end{equation}
Moreover, we can rewrite each difference term as a weighted sum of the $N$ different ensemble predictions:
\begin{equation}\small
\begin{split}
    D_i &= \frac{1}{\sqrt{N}}\left(Q_{\phi_i}^{\pi} - \frac{1}{N}\sum_{i=1}^N Q_{\phi_i}^{\pi}\right) \\
    &= \frac{1}{\sqrt{N}}\left(\frac{N-1}{N}Q_{\phi_i}^{\pi} - \sum_{j\neq i}\frac{Q_{\phi_j}}{N}\right).
\end{split}
\end{equation}
From the independence assumption of the ensemble models, it follows that each $D_i$ is also a Gaussian random variable with moments given by the basic properties for the sums of independent random variables:
\begin{equation}\small
    \begin{split}
    \E[D]  &= \mu_D = 0,\\
    var(D)  &= \sigma_D^2 = \frac{1}{N}\left(\frac{(N-1)^2}{N^2}\sigma_Q^2 + \frac{N-1}{N^2}\sigma_Q^2\right) \\
    &= \frac{N-1}{N^2}\sigma_Q^2.
    \end{split}
\end{equation}
Hence, by simply scaling the population standard deviation, we can express it as a sum of standard normal random variables: 
\begin{equation}\small
    \begin{split}
    \frac{N}{\sigma_Q\sqrt{N-1}}s &= \sqrt{\sum_{i=1}^N \left(\frac{N}{\sigma_Q\sqrt{N-1}}D_i\right)^2} \\
    &= \sqrt{\sum_{i=1}^N Z_i^2},\quad \textit{where} \quad Z\sim N(0, 1).
    \end{split}
\end{equation}
This enables to relate $s$ to a \textit{Chi} distribution \citep{chidist} with parameter $N$:
$$\frac{N}{\sigma_Q\sqrt{N-1}}s \sim \chi_N.$$ 
Consequently, the expected value of $p_\beta^s$ can be obtained from scaling the expected value of a \textit{Chi} distribution:
\begin{equation}\small
    \begin{split}
    \E[p_\beta^s] &= \beta \times \E[s]=\frac{\beta \sqrt{N-1}}{N}\E[\chi_N]\sigma_Q \\
    &=
    \frac{\beta\sqrt{N-1}}{N}\times\frac{\sqrt{2}\Gamma\left(\frac{N+1}{2}\right)}{\Gamma\left(\frac{N}{2}\right)}\sigma_Q.
    \end{split}
\end{equation}
Comparing the expected value of this penalty with the expected value of the uncertainty regularizer penalty from Eqn.~\ref{eq:exp_uncreg}, we note a few key facts. Both quantities are linearly proportional to the standard deviation of the action-value predictions $\sigma_Q$. However, for the population standard deviation penalty, the scale of the proportionality is dependent on the number of action-value models used to parameterize the critic. Hence, unlike for the uncertainty regularizer penalty, the critic's parameterization directly affects the expected penalization magnitude and should influence design choices regarding the parameter $\beta$.

\subsection{Relationship with Clipped Double Q-Learning}

We can rewrite the ubiquitous clipped double Q-learning penalization practice \citep{td3} using the uncertainty regularizer penalty. In particular, we can specify the clipped double Q-learning action-value targets from Eqn.~\ref{eq:clippeddoubleQ} in terms of the difference between the mean action-value prediction and a penalty:
\begin{equation}\small
    \begin{split}
    \hat{Q}_{\phi_{min}}(s, a)^\pi&=\min\left(Q_{\phi_1}^\pi(s, a), Q_{\phi_2}^\pi(s, a)\right) \\
    &= \bar{Q}_{\phi}^\pi(s, a) - p_{min}(s, a, \phi, \theta),\\
    \textit{where}&\quad \bar{Q}_{\phi}^\pi = \frac{1}{N}\sum_{i=1}^N Q_{\phi_i}^{\pi},\\
    p_{min} = \min ( &Q_{\phi_1}^\pi, Q_{\phi_2}^\pi ) -  \bar{Q}_{\phi}^\pi = \frac{1}{2}|Q_{\phi_1}^\pi - Q_{\phi_2}^\pi|.
    \end{split}
\end{equation}
Similarly, for $N=2$ our uncertainty regularizer reduces to:
\begin{equation}\small
    \begin{split}
    p_\beta=\frac{\beta}{2}\left(|Q_{\phi_1}^{\pi} - Q_{\phi_2}^{\pi}|+|Q_{\phi_2}^{\pi} - Q_{\phi_1}^{\pi}|\right)=\beta|Q_{\phi_1}^{\pi} - Q_{\phi_2}^{\pi}|.
    \end{split}
\end{equation}
Thus, $p_{min}=p_\beta$ for $\beta=0.5$.
As shown earlier in this section, the expected magnitude of the uncertainty regularizer penalty is not dependent on the number of action-value functions. Hence, our penalty allows us to extend the expected regularization induced by clipped double Q-learning for $N>2$, simply fixing $\beta=0.5$. This would not be possible using the population standard deviation penalty since its expected magnitude is dependent on the number of action-value functions employed.   
\section{Algorithmic and Experimental Specifications}
\label{app:params}

In this section, we provide details regarding the algorithm specification and experimental results from the main text. Our approach was to implement GPL as a general plug-in addition to existing algorithms, matching the complexity and using the same hyper-parameters of the competing modern baselines, without additional tuning. We exemplify our implementation on top of SAC in Algorithm~\ref{alg:GPLsac}, below. We also provide comprehensive ablation studies analyzing each of our design choices in Appendix~\ref{app:ext_emp_ana}. For further information about our efficient implementation, please refer to the shared code. 

\floatname{algorithm}{Procedure}
\begin{algorithm}[h]
    \small
    \caption{\textit{Uncertainty Regularization}}
    \label{alg:GPLsac_unc_reg}
    \begin{algorithmic}
    \STATE \textbf{input} $s, a, \phi, \theta, \beta$
    \STATE $p_\beta \gets \frac{\beta}{N^2-N}\sum_{i=1}^N \sum_{j\neq i}^N |Q_{\phi_i}^{\pi}(s, a) - Q_{\phi_j}^{\pi}(s, a)|$ 
    \STATE $\bar{Q}_{\phi}^\pi \gets \frac{1}{N}\sum_{i=1}^N Q_{\phi_i}^{\pi}(s, a)$
    \STATE \textbf{return} $\bar{Q}_{\phi}^\pi - p_\beta$
    
    \end{algorithmic}
\end{algorithm}

\floatname{algorithm}{Algorithm}
\begin{algorithm}[h]
    \small
    \caption{\textit{GPL-SAC Training Loop}}
    \label{alg:GPLsac}
    \begin{algorithmic}
    \STATE \textbf{input} $N, \textit{UTD}, \rho, \gamma, \hat{H}$
    \STATE \textbf{init} $\pi_\theta, \{Q_{\phi_i}^{\pi}\}^N_1, \{Q_{\phi'_i}^{\pi}\}^N_1, \mathbf{\beta\gets 0.5}, \alpha\gets 0.0, D \gets \emptyset$
    \LOOP
        \STATE Observe $s$, execute $a\sim \pi_{\theta}(s)$, collect $s', r$
        \STATE Store $D \gets D \cup (s, a, s', r)$
        \FOR{$j = 1, 2, ..., \textit{UTD}$}
        \STATE Sample minibatch $\{(s, a, s', r)\}\in D$, $a'\sim \pi(s')$
            \STATE $\hat{Q}_{\phi'}^\pi(s', a') \gets \mathbf{\textbf{Procedure 1}(s', a', \phi', \theta, \beta)}$
            \STATE $y\gets r + \gamma\left(\hat{Q}_{\phi'}^\pi(s', a') - \alpha \log \pi(a'|s')\right)$ \COMMENT{TD-targets}
            \FOR{$i = 1, 2, ..., \textit{N}$}
                \STATE $e_i\gets Q_{\phi_i}^{\pi} - y$ \COMMENT{TD-errors for the $i^{th}$ critic}
                \STATE Optimize $J(\phi_i)=e_i^2$ \COMMENT{TD-learning}
                \STATE Update $\phi'_i\gets\rho\phi'_i + (1-\rho)\phi_i$ \COMMENT{Target critic update}
            \ENDFOR
            \ENDFOR
        \STATE \textbf{Optimize $J(\beta) = \beta\times \sum_{i=1}^N e_i$ \COMMENT{\textit{dual TD-learning}}}
        \STATE Sample $a\sim \pi(s)$
        \STATE $\hat{Q}_{\phi}^\pi \gets \mathbf{\textbf{Procedure 1}(s, a, \phi, \theta, \beta)}$
        \STATE Optimize $J(\theta) = - \hat{Q}_{\phi}^{\pi} + \alpha \log \pi(a|s)$ \COMMENT{Policy learning}
        \STATE Optimize $J(\alpha) = \alpha(- \log \pi(a|s) - \hat{H})$ \COMMENT{Entropy learning}
    \ENDLOOP
    
    \end{algorithmic}
\end{algorithm}

\subsection{Empirical Bias Estimation}

\label{app:subsection:empirical_bias_estimation_desc}
In Section~\ref{sec:method}, we record the evolution of the target action-value bias throughout the RL process by running experiments with two different extensions to the SAC algorithm on three OpenAI Gym environments. The first extension makes use of unpenalized optimistic target action-values, while the second extension makes use of a penalty with a magnitude equivalent to the one induced by the clipped double Q-learning targets. Both are implemented through the uncertainty regularizer with a fixed $\beta$, as derived in Appendix~\ref{app:pen}. Moreover, both extensions use an update-to-data ratio of twenty and $N=4$ action-value models to parameterize the critic. The rest of the hyper-parameters follow our implementation of GPL-SAC. We record estimates of the action-value bias by collecting transitions from ten evaluation episodes every 1000 environment steps. We obtain each estimate by taking the average difference between the observed discounted returns in the evaluation episodes and the discounted returns from the critic's predictions. In particular, we correct the critic's raw predictions by subtracting the discounted log probabilities of the performed actions, accounting for SAC's modified action-value objective.

\subsection{Proprioceptive Observations Experiments}

\begin{table*}[t]
\centering

\begin{minipage}[t]{0.48\linewidth}

\caption{Hyper-parameters used for GPL-SAC} 
\label{tab:gplsac_hyper}
\vskip 0.15in
\tabcolsep=0.08cm
\begin{center}
\adjustbox{max width=0.98\columnwidth}{
\begin{tabular}{@{}ll@{}}
\toprule
\multicolumn{2}{c}{SAC hyper-parameters}                                                                                                                                                                                    \\ \midrule
Replay data buffer size                        & $1000000$                                                                                                                                                                  \\
Batch size                                     & $256$                                                                                                                                                                     \\
Minimum data before training                   & $5000$                                                                                                                                                                    \\
Random exploration steps                       & $5000$                                                                                                                                                                    \\
Optimizer                                      & \textit{Adam} \citep{adam}                                                                                                                                                \\
Policy/critic learning rate                    & $0.0003$                                                                                                                                                                  \\
Policy/critic $\beta_1$                        & $0.9$                                                                                                                                                                     \\
Critic UTD ratio                               & $20$                                                                                                                                                                      \\
Policy UTD ratio                               & $1$                                                                                                                                                                       \\
Discount $\gamma$                              & $0.99$                                                                                                                                                                    \\
Polyak coefficient $\rho$                      & $0.995$                                                                                                                                                                   \\
Hidden dimensionality                          & $256$                                                                                                                                                                     \\
Nonlinearity                                   & ReLU                                                                                                                                                                      \\
Initial entropy coefficient $\alpha$           & $1$                                                                                                                                                                       \\
Entropy coefficient learning rate              & $0.0001$                                                                                                                                                                  \\
Entropy coefficient $\beta_1$                  & $0.5$                                                                                                                                                                     \\
Policy target entropy $\hat{H}$                          & \begin{tabular}[c]{@{}l@{}}\textit{Hopper} : -1, \textit{HalfCheetah}: -3, \\ \textit{Walker2d}: -3, \textit{Ant}: -4, \textit{Humanoid}: -2 
\end{tabular} \\ \midrule
\multicolumn{2}{c}{GPL hyper-parameters}                                                                                                                                                                                    \\ \midrule
Initial uncertainty regularizer weight $\beta$ & $0.5$                                                                                                                                                                     \\
Uncertainty regularizer learning rate          & $0.1$                                                                                                                                                                     \\
Uncertainty regularizer $\beta_1$              & $0.5$                                                                                                                                                                     \\ \bottomrule
\end{tabular}}
\end{center}

\end{minipage}\hfill%
\begin{minipage}[t]{0.48\linewidth}
\caption{Hyper-parameters used for GPL-DrQ} 
\label{tab:gpldrq_hyper}
\vskip 0.15in
\tabcolsep=0.08cm
\begin{center}
\adjustbox{max width=0.98\columnwidth}{
\begin{tabular}{@{}ll@{}}
\toprule
\multicolumn{2}{c}{DrQv2 hyper-parameters}                                                       \\ \midrule
Replay data buffer size                         & $1000000$ ($100000$ for \textit{Quadruped run})      \\
Batch size                                      & $256$ ($512$ for \textit{Walker run})                \\
Minimum data before training                    & $4000$                                        \\
Random exploration steps                        & $2000$                                        \\
Optimizer                                       & \textit{Adam} \citep{adam}                    \\
Policy/critic learning rate                     & $0.0001$                                      \\
Policy/critic $\beta_1$                         & $0.9$                                         \\
Critic UTD ratio                                & $0.5$                                         \\
Policy UTD ratio                                & $0.5$                                         \\
Discount $\gamma$                               & $0.99$                                        \\
Polyak coefficient $\rho$                       & $0.99$                                        \\
\textit{N}-step returns                         & $3$ ($1$ for \textit{Walker run})                    \\
Hidden dimensionality                           & $1024$                                        \\
Feature dimensionality                          & $50$                                          \\
Nonlinearity                                    & ReLU                                          \\
Initial entropy coefficient $\alpha$            & $1$                                           \\
Exploration stddev. clip                        & $0.3$                                         \\
Exploration stddev. schedule                    & linear: $1 \rightarrow 0.1$ in $500000$ steps \\ \midrule
\multicolumn{2}{c}{GPL hyper-parameters}                                                         \\ \midrule
Initial uncertainty regularizer weight $\beta$  & $0.5$                                         \\
Uncertainty regularizer learning rate           & $0.1$                                         \\
Uncertainty regularizer $\beta_1$               & $0.5$                                         \\ \midrule
\multicolumn{2}{c}{Pessimism annealing hyper-parameters}                                         \\ \midrule
Optimistic shift value $\lambda_{opt}$ schedule & linear: $0.5 \rightarrow 0$ in $500000$ steps \\ \bottomrule
\end{tabular}}
\end{center}

\end{minipage}

\end{table*}

\begin{figure}
  \centering
  \includegraphics[width=0.75\columnwidth]{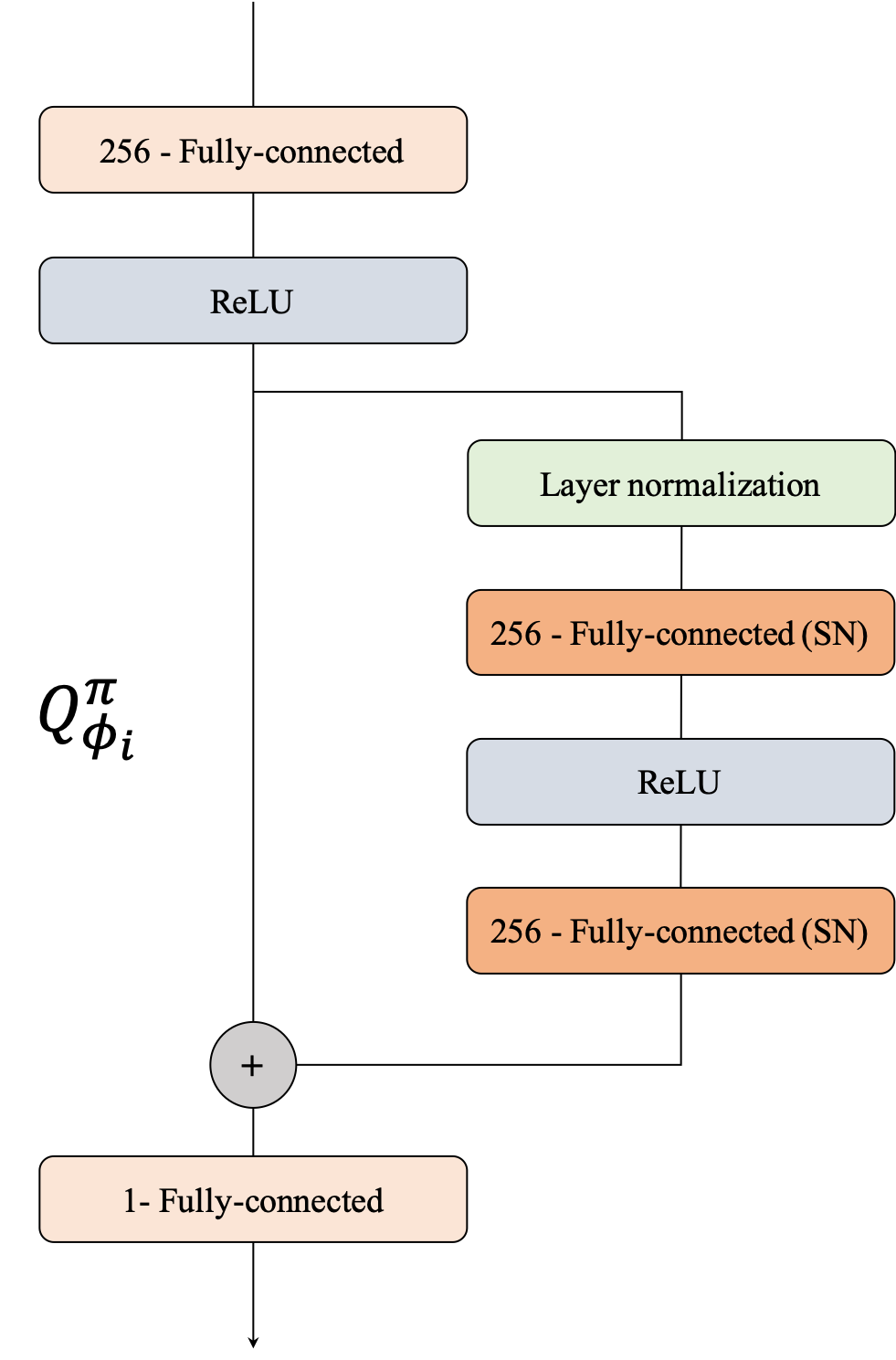}
  \caption{Schematic representation of the modern model architecture used to parameterize the action-value functions in our implementation of GPL-SAC. We make use of a single hidden residual block where the fully-connected layers are regularized via spectral normalization.}
  \label{figure:modern-arch}
\end{figure}

\textbf{Hyper-parameters.} We provide a list of the utilized hyper-parameters for GPL-SAC in Table~\ref{tab:gplsac_hyper}. The majority of the chosen hyper-parameters follow closely the seminal SAC papers \citep{sac, sac-alg}, with some exceptions to account for modern implementation advancements. For instance, consistently with recent literature, we employ increased model-ensembles and update-to-data ratios which appear to be crucial components for sample-efficiency in RL (see \citep{sunrise, redq, mbpo} and also Appendix~\ref{appsub:ens}-\ref{appsub:utd}). To be consistent with the compared baselines we use ten action-value models and a UTD of twenty. Additionally, we employ a simplified version of the modern architecture from \citet{deeper-deep-RL} to parameterize the action-value models, as depicted in Figure~\ref{figure:modern-arch}. This architecture follows many of the practices introduced by recent work to stabilize transformer training \citep{preln-trans}. Specifically, we employ a single hidden residual block with layer normalization \citep{layernorm} followed by two fully-connected layers regularized with spectral normalization \citep{sngan}. While we observed relatively small performance improvements for this choice over a simpler 3-layer fully-connected architecture (see Appendix~\ref{appsub:ens}), we also observed only marginal extra computational demand, with our GPL-SAC still running considerably faster than REDQ (see Appendix~\ref{appsub:comp_scaling}). Throughout all architectures, we keep the hidden dimensionality fixed to 256. We initialize the uncertainty regularizer with $\beta=0.5$ to reflect the penalization magnitude of clipped double Q-learning (see Appendix~\ref{app:pen}). Following \citet{sac-alg}, the entropy coefficient $\alpha$ is adaptively updated at each training iteration based on a policy target entropy $\hat{H}$. In particular, this follows a dual optimization procedure to keep the estimated average policy entropy close to $\hat{H}$, as described in Line 17 of Algorithm~\ref{alg:GPLsac}.  We follow the values of $\hat{H}$ from \citet{mbpo}. We learn $\beta$ using dual TD-learning with the same optimizer used for adjusting the value of $\alpha$. 

\textbf{Baseline results.} For the continuous control experiments from proprioceptive states, we employ different baselines to ground our results and provide a comparison with current state-of-the-art algorithms. The reported results for REDQ come from re-running \citet{redq}'s original implementation with the provided hyper-parameters. The reported results for MBPO come instead from the original paper, as publicly shared by \citet{mbpo}. The reported results for SAC-20 come from running the same base implementation as GPL-SAC, with a few differences in the listed hyper-parameters. Namely, SAC-20 uses an ensemble of $N=2$ action-value models with the classical 3-layer fully-connected architecture from \citet{sac-alg} and uses an uncertainty regularizer penalty with a \textit{fixed} parameter $\beta=0.5$.

\subsection{Pixel Observations Experiments}

\textbf{Hyper-parameters.} We provide a list of the utilized hyper-parameters for GPL-DrQ in Table~\ref{tab:gpldrq_hyper}. All the hyper-parameters shared with DrQv2 follow the values provided by \citet{drqv2}, while the uncertainty regularizer and optimizer for $\beta$ follow the same specifications as in GPL-SAC. Additionally, GPL-DRQ-Anneal linearly decays the optimistic shift value $\lambda_{opt}$ from 0.5 down to 0.0 with the same frequency as the exploration noise's standard deviation.

\textbf{Baseline results.} For the experiments on the DeepMind Control Suite from pixel observations, we compare our extensions with the base DrQv2 algorithm. Since the provided results for DrQv2 had inconsistent numbers of repetitions, we recollected the results by running the experiments with \citet{drqv2}'s original implementation. However, in our evaluation, we observed higher variances in the performance for some of the considered tasks than what \citet{drqv2} reported. We shared this inconsistency with DrQv2's authors, and they confirmed the validity of our empirical findings after recollecting the results themselves.

\subsection{Statistical Significance Analysis}

We evaluate the statistical significance of all considered algorithms utilizing the evaluation protocol proposed by~\citet{precipice-rliable}. For the continuous control experiments from proprioceptive states, we normalize each task's final returns using the performance obtained by SAC after 3M steps, i.e., collecting 30 times the amount of experience, enough to reach convergence on most examined problems. The DeepMind Control Suite's tasks already provide returns scaled within 0 and 1000, which we naturally use for normalization. When computing stratified bootstrap confidence intervals \citep{bootstrap-cis}, we use 2000 samples for the probability of improvements and 50000 samples for all other metrics (as suggested by~\citet{precipice-rliable}).

\section{Main Performance Results Extension}
\label{app:res_vis}

\subsection{OpenAI Gym Unscaled Results after 100k Experience Steps}

\label{app:res_vis_mj}
\begin{table}[H]
\caption{Results for the experiments on the OpenAI Gym suite after collecting 100K experience steps}
\label{tab:100k}
\vspace{-10pt}
\tabcolsep=0.03cm
\begin{center}
\adjustbox{max width=0.98\linewidth}{
\begin{tabular}{@{}lcccc@{}}
\toprule
\textbf{Algorithm / Task}   & SAC-20         & REDQ & MBPO & GPL-SAC (Ours)          \\ \midrule
\textit{Hopper-v2}          & $2694\pm 902$  & $3007\pm 471$                      & $3262\pm 197$                      & $\mathbf{3386\pm 92}$   \\
\textit{Halfcheetah-v2}     & $5822\pm 728$  & $5625\pm 431$                      & $9501\pm 331$                      & $\mathbf{9685\pm 658}$  \\
\textit{Walker2d-v2}        & $2101\pm 876$  & $1937\pm 968$                      & $3377\pm 529$                      & $\mathbf{3662\pm 360}$  \\
\textit{Ant-v2}             & $482\pm 101$   & $3160\pm 1213$                     & $1624\pm 447$                      & $\mathbf{4933\pm 333}$  \\
\textit{Humanoid-v2}        & $493\pm 94$    & $1460\pm 689$                      & $555\pm 61$                        & $\mathbf{5155\pm 191}$  \\ \midrule
\textit{Normalized average} & $0.35\pm 0.28$ & $0.48\pm 0.24$                     & $0.52\pm 0.30$                     & $\mathbf{0.81\pm 0.12}$ \\ \bottomrule
\end{tabular}}
\end{center}
\vspace{-5pt}
\end{table}

\begin{figure*}
  \centering
  \includegraphics[width=\linewidth]{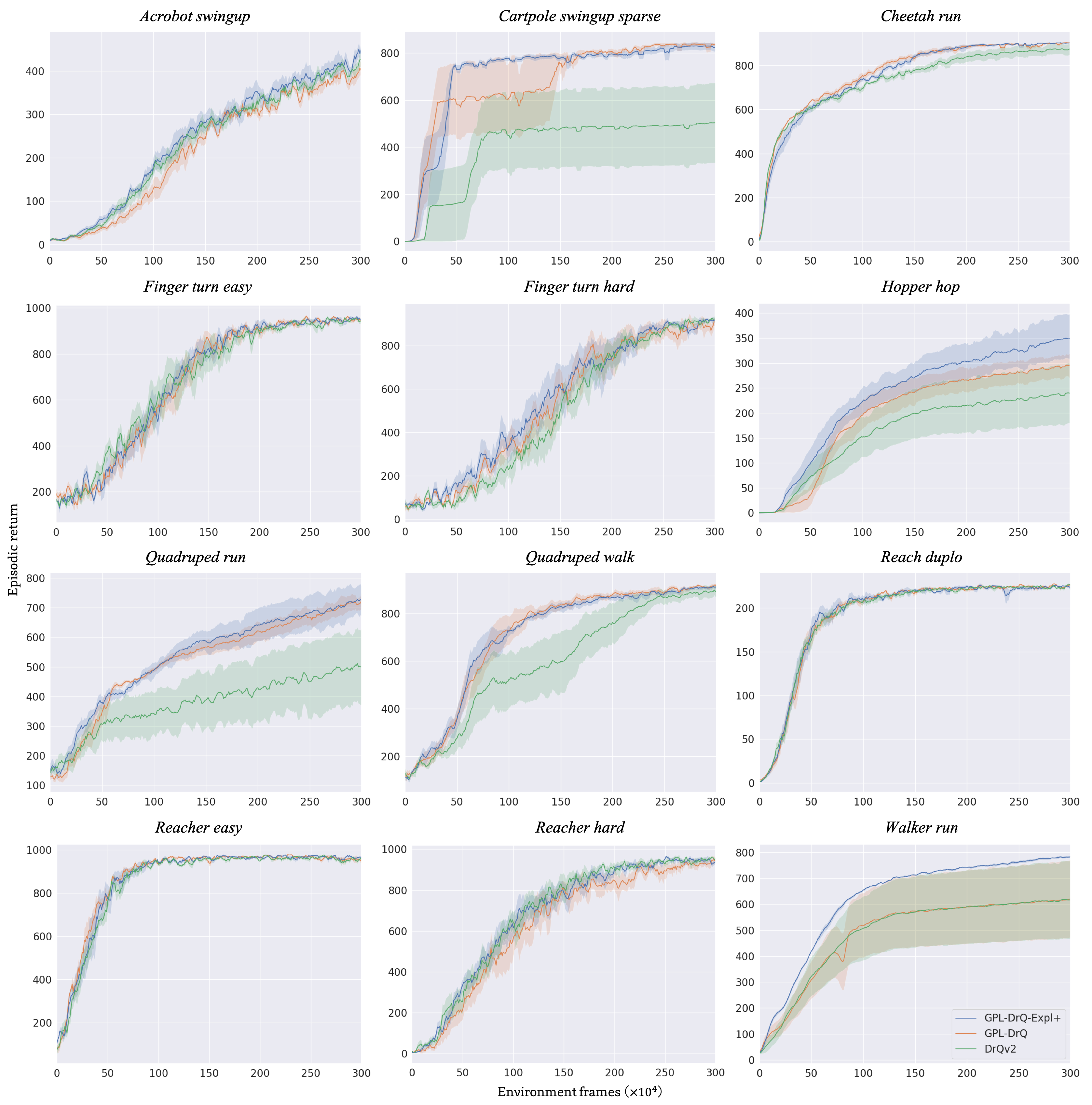}
  \caption{Performance curves for the environments of the DeepMind Control Suite. We report both mean and standard deviation of the episodic returns over five random seeds.} 
  \label{figure:dmc_res}
\end{figure*}

In Table~\ref{tab:100k} we compare the performances of the examined algorithms after the milestone of collecting 100K environment steps.  We believe this to be an appropriate comparison point for recent algorithms, given their relative sample-efficiency improvements in the OpenAI Gym tasks. For each run, we record the average returns from 25 evaluation episodes collected in the preceding 5000 environment steps. These results highlight once again the superior sample efficiency of GPL-SAC, pushing the performance boundary of off-policy methods. GPL-SAC is also more consistent than the considered baselines in most tasks, as shown by the lower relative standard deviations. As mentioned in Section~\ref{sec:experiments} of the main text, we attribute the performance gap mainly to the ability of GPL to prevent the accumulation of overestimation bias without requiring overly pessimistic targets. 
In particular, the principal identified downsides of overly pessimistic targets are two-fold.
Firstly, they slow down reward propagation and introduce further errors in the TD-targets, potentially leading to suboptimal convergence of the relative action-value functions \citep{discor}. Secondly, they induce overly conservative policies that prefer low-uncertainty behavior, hindering exploration in stochastic environments \citep{optadv-OAC}.

\subsection{DeepMind Control Suite Extended Results}

\label{app:res_vis_dmc}


\begin{table*}[t]
\caption{Extended version of Table~\ref{tab:gpldrq} with detailed per-environment results for the DeepMind Control Suite after collecting 1.5M and 3M experience step.} \label{tab:gpldrq_extended}
\vskip 0.1in
\tabcolsep=0.07cm
\begin{center}
\adjustbox{max width=0.95\linewidth}{
\begin{tabular}{@{}lcccccccccc@{}}
\toprule
\textbf{Evaluation milestone}          & \multicolumn{5}{c}{3.0M frames}                                          & \multicolumn{5}{c}{1.5M frames}                                            \\ \midrule
\textbf{Task\textbackslash{}Algorithm} & GPL-DrQ-Anneal    & GPL-DrQ         & DrQv2           & CURL    & SAC     & GPL-DrQ-Anneal    & GPL-DrQ          & DrQv2            & CURL    & SAC     \\ \cmidrule(lr){2-6} \cmidrule(lr){7-11}
\textit{Acrobot swingup}               & \textbf{446±33}  & 393±29          & 414±34          & 6±3     & 10±5    & \textbf{283±49}  & 246±31           & \textbf{277±39}  & 5±2     & 7±3     \\
\textit{Cartpole swingup sparse}       & 824±33           & \textbf{837±15} & 503±411         & 530±350 & 169±291 & \textbf{780±25}  & 740±123          & 475±388          & 498±327 & 135±268 \\
\textit{Cheetah run}                   & \textbf{902±2}   & \textbf{903±1}  & 871±52          & 586±89  & 7±7     & \textbf{830±25}  & \textbf{837±29}  & 771±24           & 506±86  & 9±6     \\
\textit{Finger turn easy}              & \textbf{952±15}  & \textbf{945±18} & \textbf{946±16} & 283±130 & 202±62  & 810±84           & \textbf{860±74}  & 794±159          & 279±97  & 157±45  \\
\textit{Finger turn hard}              & \textbf{918±24}  & 893±59          & \textbf{924±17} & 180±88  & 88±51   & \textbf{615±233} & \textbf{587±238} & 484±156          & 240±150 & 76±46   \\
\textit{Hopper hop}                    & \textbf{349±90}  & 296±52          & 240±123         & 222±133 & 0±0     & \textbf{252±76}  & \textbf{242±56}  & 198±101          & 186±128 & 0±0     \\
\textit{Quadruped run}                 & \textbf{725±112} & \textbf{712±51} & 504±279         & 169±87  & 57±21   & \textbf{589±71}  & \textbf{564±54}  & 385±214          & 190±88  & 56±35   \\
\textit{Quadruped walk}                & \textbf{912±16}  & \textbf{918±16} & 897±45          & 144±20  & 52±19   & \textbf{826±35}  & \textbf{834±46}  & 591±270          & 133±48  & 65±41   \\
\textit{Reach duplo}                   & \textbf{225±3}   & \textbf{226±1}  & \textbf{227±2}  & 8±8     & 1±1     & \textbf{219±4}   & \textbf{217±6}   & \textbf{219±6}   & 10±13   & 0±0     \\
\textit{Reacher easy}                  & \textbf{962±7}   & \textbf{957±12} & 952±17          & 636±176 & 89±33   & \textbf{968±8}   & 951±21           & 961±14           & 701±69  & 68±41   \\
\textit{Reacher hard}                  & 933±18           & 946±37          & \textbf{957±13} & 612±301 & 11±12   & \textbf{798±115} & \textbf{790±104} & \textbf{813±122} & 508±152 & 5±10    \\
\textit{Walker run}                    & \textbf{782±10}  & 618±298         & 617±296         & 446±223 & 26±4    & \textbf{713±5}   & 574±275          & 569±273          & 378±233 & 26±4    \\ \midrule
\textbf{Average score}                 & \textbf{744.09}  & 720.29          & 670.95          & 318.38  & 59.38   & \textbf{640.20}  & 620.24           & 544.67           & 302.74  & 50.34   \\ \midrule
\textbf{Top score count}               & \textbf{10/12}   & 7/12            & 4/12            & /12     & /12     & \textbf{11/12}   & 8/12             & 3/12             & /12     & /12     \\ \bottomrule
\end{tabular}
}

\end{center}
\end{table*}

In Table~\ref{tab:gpldrq_extended} we provide the detailed per-environment results for the pixel-based medium benchmark tasks from the DeepMind Control Suite after collecting 1.5M and 3M experience steps, extending the results summary provided in Table~\ref{tab:gpldrq}. For each task, we \textbf{emphasize} the top performances that are within half a standard deviation from the highest mean return. Both GPL-DrQ and GPL-DrQ-Anneal significantly improve the performance of DrQv2 \citep{drqv2} and all other baseline algorithms in the great majority of the tasks (7 and 11 out of 12, respectively). We posit that DrQv2's inconsistent performance on some tasks of the sparse reward and higher-dimensional tasks (e.g., Cartpole swingup sparse, Hopper hop, Quadruped run, Walker run...) is likely due to a lack of exploration from its overly pessimistic critic. This observed inconsistency was also observed by DrQv2's authors after running additional experiments where they re-collected the paper's original results. Incorporating GPL generally appears to resolve this issue, while pessimism annealing further aids precisely in the tasks where under-exploration is more frequent. These findings appear to to support the further validate the significance of enforcing the right levels of pessimism in off-policy RL, together with the efficacy of our new framework.

In Figure~\ref{figure:dmc_res} we provide the performance curves for the pixel-based medium benchmark tasks from the DeepMind Control Suite. These visualizations complement the results summary provided in Tables \ref{tab:gpldrq} and \ref{tab:gpldrq_extended}, further highlighting the effectiveness of GPL and the proposed pessimism annealing procedure. In accordance with our earlier analysis, our methods yield improved performance and robustness even from the very first training iterations for most of the harder exploration environments, where overly pessimistic action-value targets are expectedly more detrimental.


\begin{figure*}[t]
  \centering
  \includegraphics[width=0.8\linewidth]{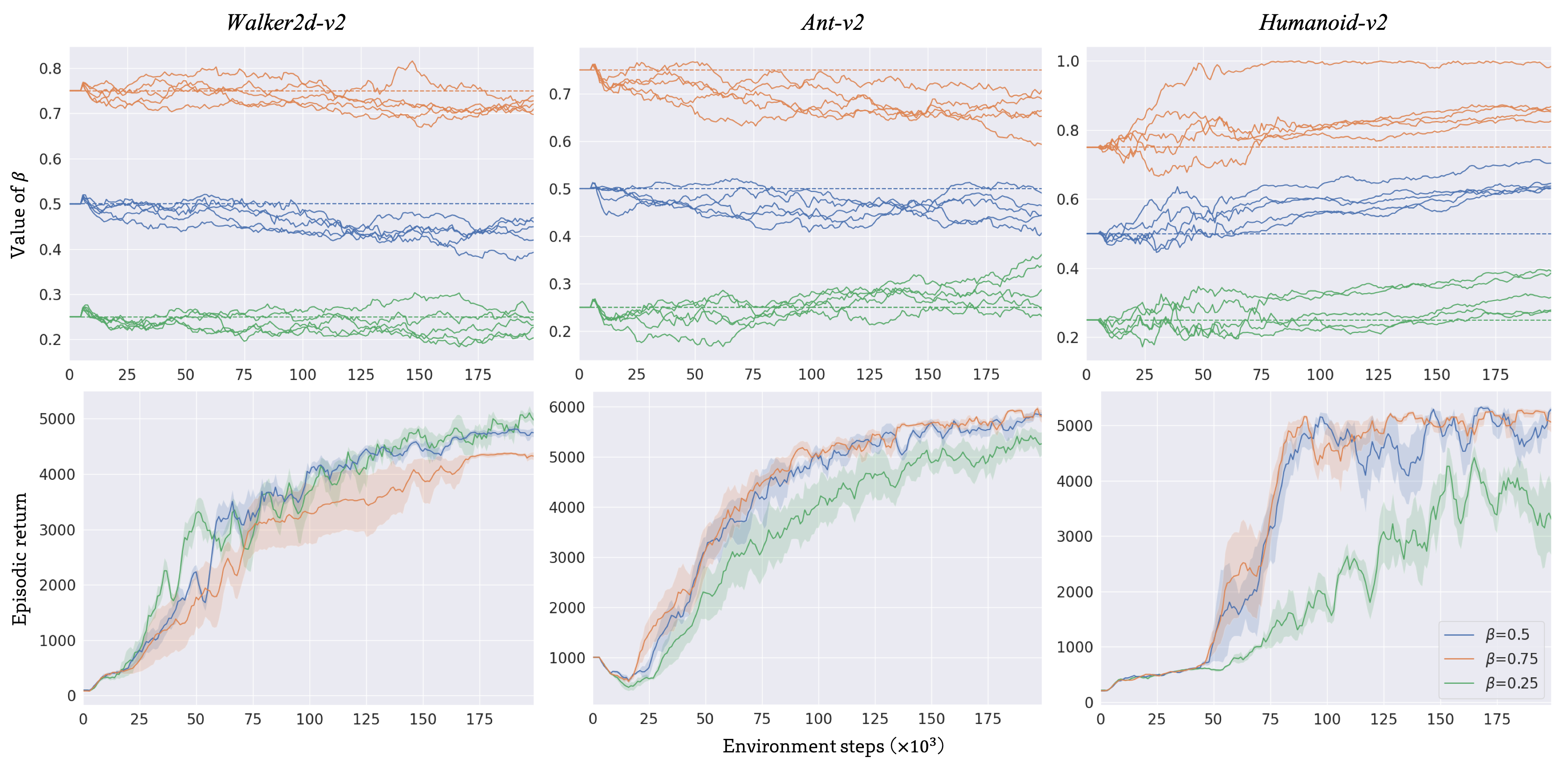}
  \caption{Analysis of the uncertainty regularizer parameter. We consider three versions of GPL-SAC with different initial values for $\beta$. We show the evolution of $\beta$ throughout optimization in each experiment (Top). We also provide the relative performance curves for each initial setting (Bottom).} 
  \label{figure:ini_of_beta}
\end{figure*}

\section{Extended Empirical Analysis}
\label{app:ext_emp_ana}

\begin{figure*}[t]
  \centering
  \includegraphics[width=0.8\linewidth]{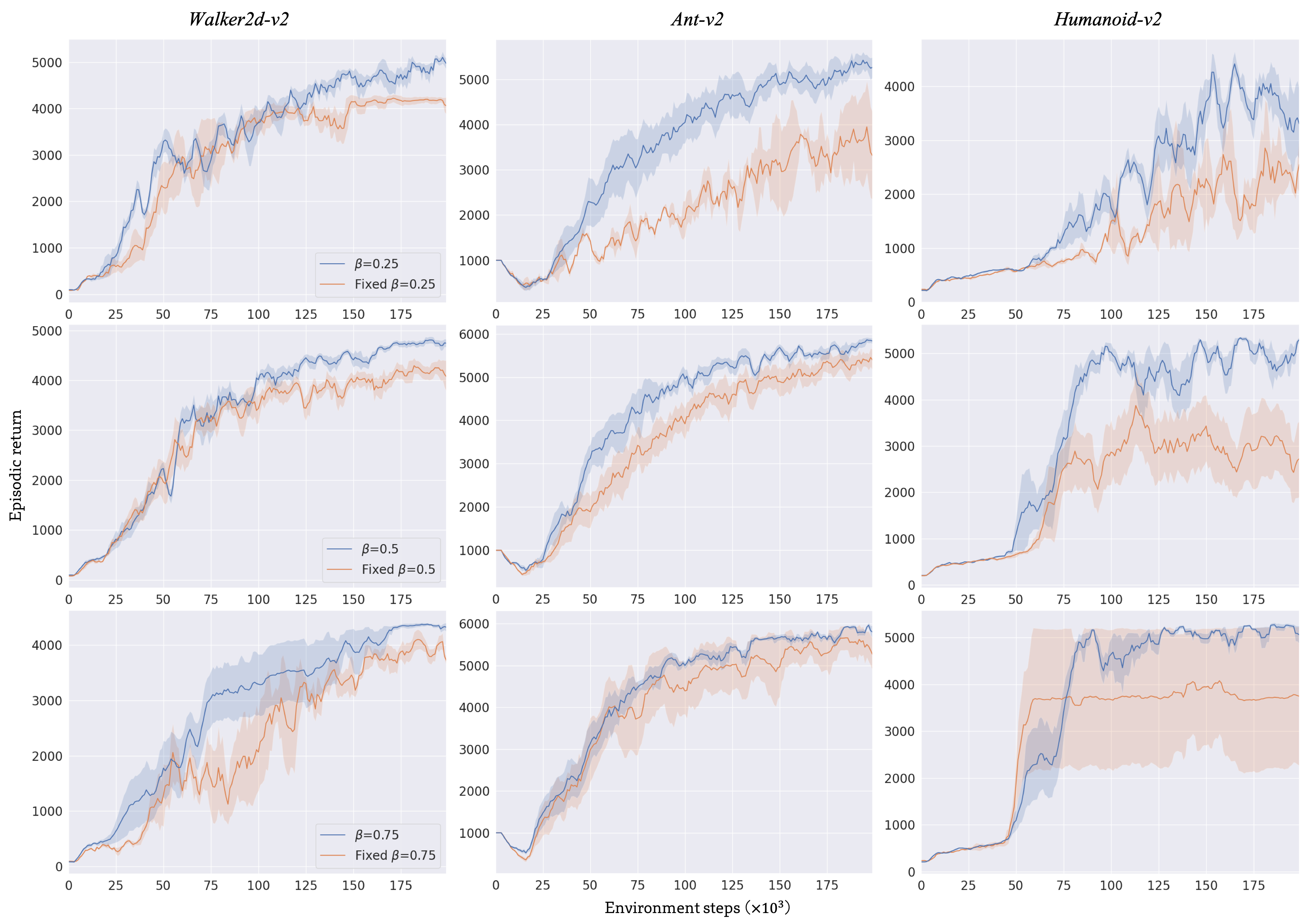}
  \caption{Performance curves showing the effects of using an uncertainty regularizer with a fixed $\beta$ in GPL-SAC, as compared to the proposed standard \textit{pessimism learning}. We show direct performance comparisons for each considered initial setting (rows) and each considered ablation task (columns).} 
  \label{figure:all_beta_perf_abl}
\end{figure*}
\begin{figure*}
  \centering
  \includegraphics[width=0.8\linewidth]{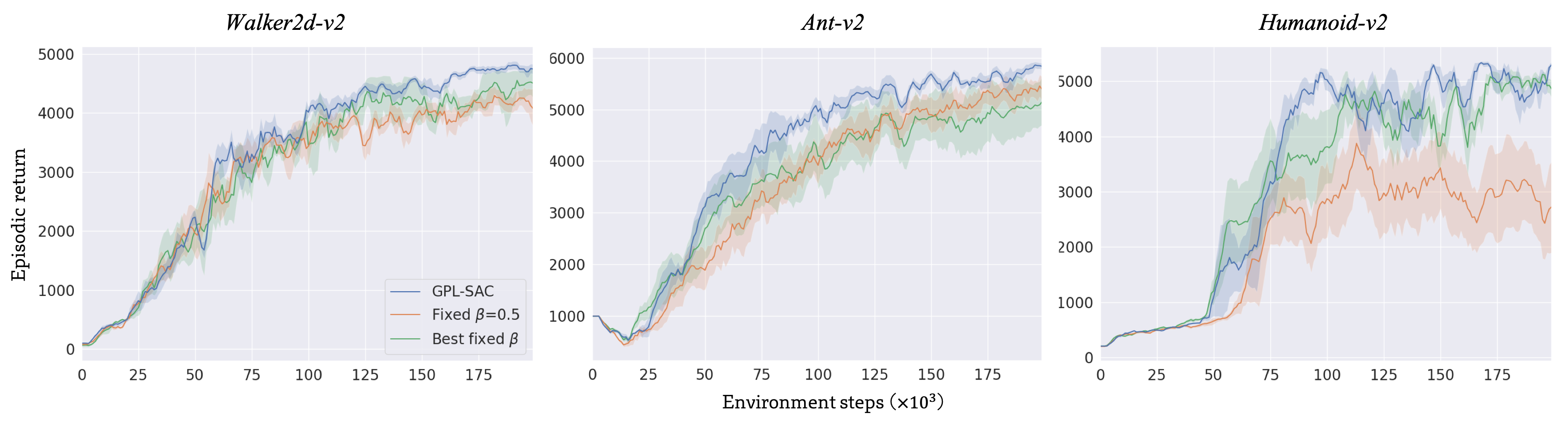}
  \caption{Performance curves comparing the average best found fixed $\beta$ in GPL-SAC, to the proposed standard \textit{pessimism learning}. We also show results using a standard fixed of $\beta=0.5$ for reference.}
  \label{figure:best_fixed_beta}
\end{figure*}

We provide an extended empirical analysis of Generalized Pessimism Learning. In particular, we study the effects of different components and parameters on the stability and efficiency of our GPL-SAC algorithm. We focus our analysis on a representative subset of the considered OpenAI Gym tasks. We follow the same experimental protocol as described in Section~\ref{sec:experiments} of the main text.

\subsection{Pessimism Learning Analysis}
\label{appsub:unc_reg}
\label{ext:unc_reg}

\textbf{Recorded GPL bias.} We provide an extension the results in Figure~\ref{figure:mujoco_bias} which analyze periodic recordings of the empirical action-value bias by comparing the actual and estimated discounted returns of a SAC agent with different fixed bias-counteraction penalties (as described in App. \ref{app:subsection:empirical_bias_estimation_desc}). 

\begin{figure}[H]
  \centering
  \includegraphics[width=0.8\columnwidth]{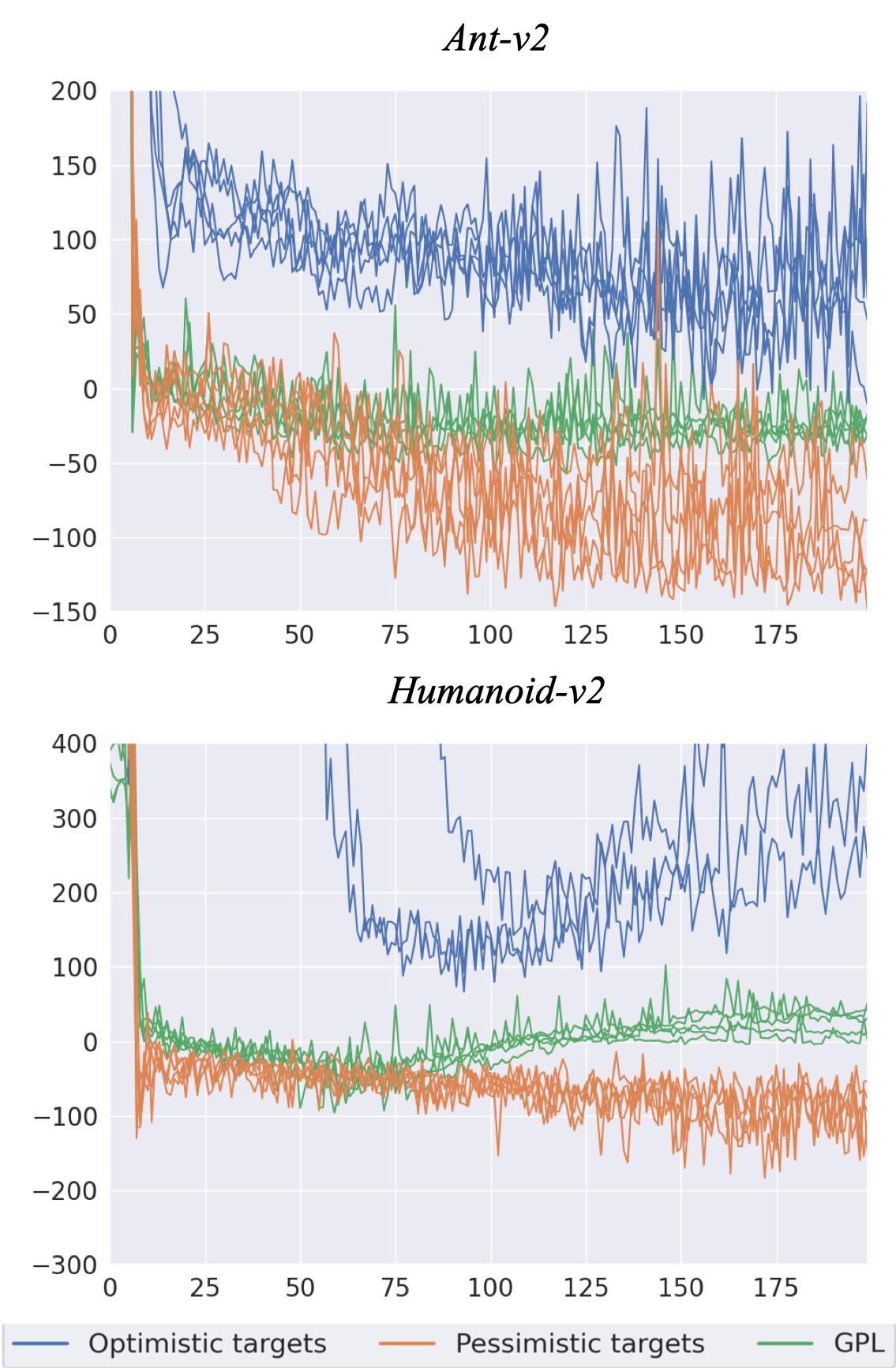}
  \caption{Recorded estimated bias training SAC agents with either fixed (optimistic/pessimistic) or tuned (GPL) action-value targets.} 
  \label{figure:bias_pess_v_opt_v_gpl}
\end{figure}

In Figure~\ref{figure:bias_pess_v_opt_v_gpl}, we compare the estimates using the described optimistic and pessimistic targets with \textit{adaptive} targets learned throughout the training procedure by using GPL. As observed in Section~\ref{subsec:dual_td}, for both the pessimistic and optimistic targets the bias in the predicted action-values varies considerably across different environments, agents, training stages, and random seeds. In contrast, regulating bias through GPL consistently leads to bias much closer to zero, with considerably smaller fluctuations.

\textbf{$\beta$ initialization.} In all main experiments, we initialize the uncertainty regularizer with $\beta=0.5$. This setting yields the same initial expected regularization magnitude as the ubiquitous clipped double Q-learning method. Hence, to better understand the optimization properties of GPL, we consider instantiating GPL-SAC with three different initial values for $\beta$, namely, $\beta=0.25$, $\beta=0.5$, and $\beta=0.75$. For these settings, we record both the evolution of $\beta$ throughout learning and the performance of the resulting algorithms. Finally, we also also directly compare performances when \textit{ablating} the pessimism learning component of GPL-SAC, maintaining a fixed $\beta$ throughout each experiment. This analysis aims to evaluate the sensitivity of GPL to the initial value of $\beta$ and the effectiveness of the dual TD-learning procedure. In particular, this simple experimental setting should elucidate some of the properties of the actual implementation of GPL, where the initial unbiasedness assumption used to motivate dual TD-learning does not necessarily hold. Moreover, the learned values of $\beta$ should provide insights regarding the bias arising from interactions of off-policy reinforcement learning methods and the different examined tasks.

\textbf{Parameter evolution.} In Figure~\ref{figure:ini_of_beta} (Top), we show the value of $\beta$ collected throughout training in each experiment with the examined initial settings. For each task, $\beta$ appears to follow a recognizable trend of convergence towards some particular distinct range of values. This range appears to be influenced by the environment's complexity, with $\beta$ converging to lower values for \textit{Walker2d} and increasingly higher values for \textit{Ant} and \textit{Humanoid}, respectively. However, $\beta$ appears to adapt rather slowly, seldom reaching stability for distant initialization values in the examined experience regime. Our intuition is that this phenomenon is mainly due to two characteristics of pessimism learning. 
The first characteristic is that bias in the target action-value predictions can arise simply due to the stochasticity of the RL process dynamics for any value of $\beta$.
Hence, there will always be a stochastic component from the dual-TD learning signal introducing noise to $\beta$'s optimization.
The second characteristic is that, in practice, dual TD-learning occurs at a slower rate than TD-learning itself. Hence, as discussed in Section~\ref{sec:method}, given an imperfect initialization of $\beta$, part of the target bias might leak into the online action-value models in the first iterations of training. While the iterative interactions between dual TD-learning and TD-learning still pushes the value of $\beta$ to mitigate bias, consistently with our inituition, training takes longer to reach expectedly unbiased targets from a dimmed-down signal in the `unbiasing' learning direction. This suggests that despite empirical effectiveness, dual TD-learning might further benefit from improvements to the dual-objective that take into account potential \textit{momentum} terms, to counteract learning slowdowns from this phenomenon.

\textbf{Performance results.} In Figure~\ref{figure:ini_of_beta} (Bottom), we show the performance curves for the examined initial settings. While GPL-SAC recovers strong performance for most tested initializations, we can see robustness improve when we initialize $\beta$ closer to its convergence range. Specifically, $\beta=0.75$ appears to work best for \textit{Humanoid}, $\beta=0.5$ for \textit{Ant}, and $\beta=0.25$ for \textit{Walker}. These results provide further evidence that one of the main factors determining optimal pessimism is task complexity, with harder tasks requiring more conservative targets for better stabilization. Thus, these results further highlight the importance of an adaptive strategy for off-policy RL to achieve proper bias counteraction in arbitrary environments.

\textbf{Pessimism performance improvements.} In Figure~\ref{figure:all_beta_perf_abl}, we show direct performance comparisons between performing pessimism learning (default) and using the uncertainty regularizer with a \textit{fixed} value of $\beta$ throughout training. Expectedly, utilizing a fixed $\beta$ appears to bring the performance of our implementation much closer to REDQ (see Section~\ref{sec:experiments}). We observe notable performance gains for all initial settings and all tasks, further validating empirical effectiveness. Moreover, we observe larger gains for higher-dimensional problems with miss-specified initial $\beta$ (e.g. Ant with $\beta=0.25$ and Humanoid), suggesting that GPL's benefit could effectively scale with the difficulty of the underlying RL problem. Furthermore, they indicate that our adaptive learning procedure can make off-policy algorithms more resilient to utilizing initial suboptimal parameters to regulate pessimism and display improved robustness. 

\textbf{Best fixed $\beta$ values.} In Figure~\ref{figure:best_fixed_beta}, we evaluate GPL-SAC with additional fixed environment-dependent values values for $\beta$. In particular, we obtained these values by averaging the final $\beta$ in the three best performing runs for each environment, from all runs in the previous experiment. These values correspond to $\beta=0.86$ for Humanoid, $\beta=0.53$ for Ant, and $\beta=0.34$ for Walker. Again, we show direct performance comparisons with performing pessimism learning, and fixing $\beta$ to its default initial value of 0.5. Overall, we find these environment-dependent values of $\beta$ work considerably better than using any of the fixed shared values considered in the previous experiment. For instance, final performance is close to the performance of our pessimism learning approach in both Walker and Humanoid environments. However, the full adaptive GPL-SAC still consistently attains the best overall results. We believe these results provide additional evidence in support of our claim that no fixed penalty can best counteract overestimation bias, given its stochasticity and non-stationarity.


\subsection{Alternative Uncertainty Regularizer Optimizations}
\label{ext:alt_opt_beta}
\label{appsub:alt_opt_beta}

\begin{figure*}[t]
  \centering
  \includegraphics[width=0.8\linewidth]{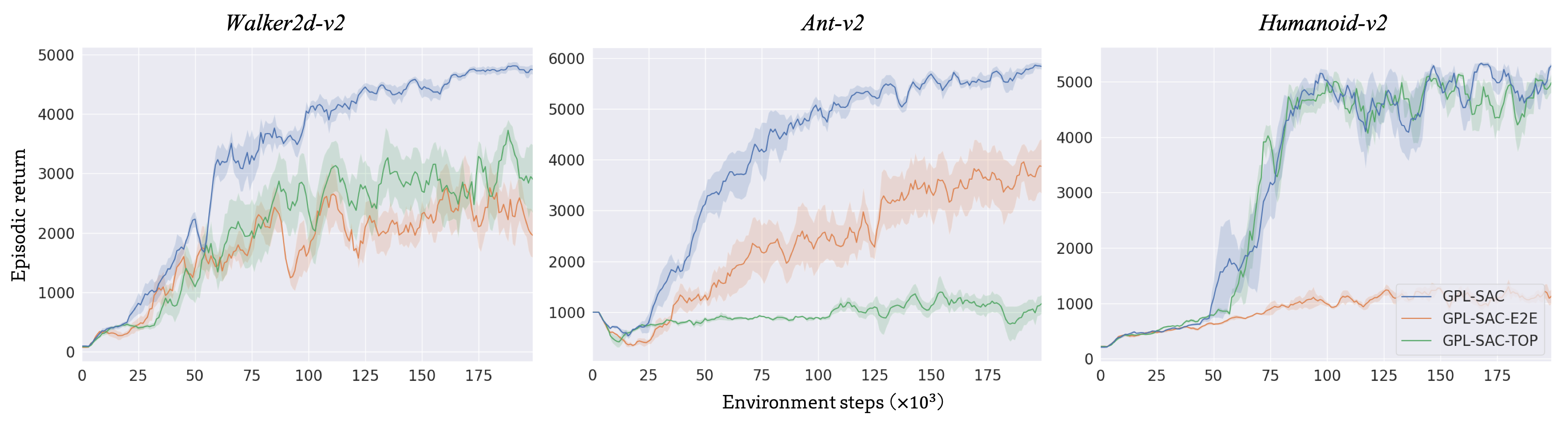}
  \caption{Performance curves showing the effects of optimizing the uncertainty regularizer with alternative strategies in GPL-SAC. We consider learning $\beta$ to minimize the estimated bias end-to-end and maximize immediate returns improvements via a bandit-based optimization from prior work.} 
  \label{figure:e2e_opt}

  \captionof{table}{Results from replacing dual TD-learning with the bandit-based optimization from prior work on a subset of the DeepMind Control environments}\label{app:tab:top_dmc}
\vskip 0.15in
\tabcolsep=0.12cm
\begin{center}
\adjustbox{max width=0.75\linewidth}{
\begin{tabular}{@{}lcccccc@{}}
\toprule
\textbf{Evaluation milestone}          & \multicolumn{3}{c}{3.0M frames}         & \multicolumn{3}{c}{1.5M frames}         \\ \midrule
\textbf{Task\textbackslash{}Algorithm} & GPL-DrQ         & GPL-DrQ-TOP & DrQv2   & GPL-DrQ         & GPL-DrQ-TOP & DrQv2   \\
Cheetah run                            & 899±4           & 898±1       & 873±60  & 843±48          & 777±50      & 792±29  \\
Hopper hop                             & 297±52          & 290±3       & 240±123 & 238±48          & 233±34      & 198±102 \\
Quadruped run                          & 710±84          & 562±129     & 523±271 & 569±54          & 517±43      & 419±204 \\
\textbf{Average score}                 & \textbf{635.41} & 583.23      & 545.16  & \textbf{550.12} & 509.21      & 470.03  \\ \bottomrule
\end{tabular}
}
\end{center}
\end{figure*}


To evaluate the relative empirical effectiveness of dual TD-learning, we implement and test two alternative optimization strategies for the uncertainty regularizer $p_{\beta}$. In particular, we replace the dual TD-learning procedure in GPL-SAC with each strategy and compare the resulting performances.

\begin{figure*}[t]
    \centering
  \includegraphics[width=0.8\linewidth]{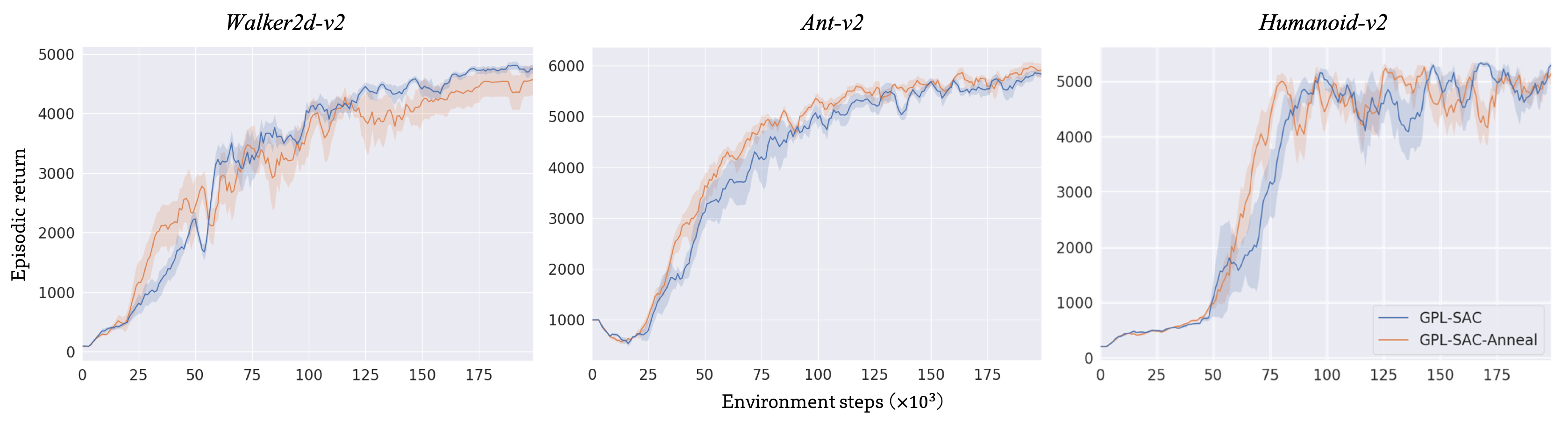}
  \caption{Performance curves showing the effects of applying pessimism annealing to GPL-SAC for the considered OpenAI Gym tasks.} 
  \label{figure:pess_ann}

  \captionof{table}{Results from doubling or halving the initial optimistic shift value $\lambda_{opt}$ of GPL-DrQ-Anneal} \label{app:tab:optsh}
\vskip 0.15in
\tabcolsep=0.08cm
\small
\begin{center}
\adjustbox{max width=0.8\linewidth}{
\begin{tabular}{lcccccccc}
\hline
\textbf{Evaluation milestone} & \multicolumn{4}{c}{3.0M frames}                                             & \multicolumn{4}{c}{1.5M frames}                                         \\ \hline
\textbf{Task\textbackslash{}Algorithm}       & $\lambda_{opt}=1$     & $\lambda_{opt}=0.5$   & $\lambda_{opt}=0.25$   & GPL-DrQ & $\lambda_{opt}=1$ & $\lambda_{opt}=0.5$   & $\lambda_{opt}=0.25$  & GPL-DrQ \\
Cheetah run                   & 899±11          & 900±9           & 901±4           & 899±4                 & 754±84      & 846±22          & 806±81          & 843±48                \\
Hopper hop                    & 323±94          & 346±93          & 358±45          & 297±52                & 241±73      & 246±72          & 282±42          & 238±48                \\
Quadruped run                 & 745±59          & 727±128         & 709±111         & 710±84                & 576±67      & 587±66          & 572±57          & 569±54                \\
\textbf{Average score}        & \textbf{655.57} & \textbf{657.80} & \textbf{655.97} & 635.41                & 523.52      & \textbf{559.42} & \textbf{553.58} & \textbf{550.12}       \\ \hline
\end{tabular}
}

\end{center}
\end{figure*}

\textbf{\textit{GPL-SAC-E2E}.} First, we consider learning $\beta$ to minimize end-to-end the squared norm of the expected target action-value bias approximation from Eqn.~\ref{approx_bias}. This optimization strategy could be seen as a more direct approach than dual TD-learning, resulting in the following optimization objective:
\begin{equation}\small
\label{eqn:e2e_bias_min}
    \begin{split}
        \argmin_\beta \E_{s, a, s}\left[B(s, a, s'|\beta)^2\right].
    \end{split}
\end{equation}
In practice, we optimize this alternative objective in place of dual TD-learning with standard gradient descent. We name the resulting algorithm \textit{GPL-SAC-E2E}. 

\textbf{\textit{GPL-SAC-TOP}.} We also consider the alternative optimization procedure from the \textit{Tactical Optimism and Pessimism (TOP)} algorithm \citep{top}. In particular, this optimization involves learning an adaptive binary controller that switches on or off a bias correction penalty based on the sample standard deviation of the critic's action-value estimates. The TOP algorithm learns this controller as a multi-armed bandit problem, using the difference in consecutive episodic returns as feedback. We transpose this framework to GPL by optimizing $\beta$ over a discrete choice of two possible values, $\beta\in\{0, \sqrt{2}\}$. With these values, the uncertainty regularizer yields the same expected penalization as the optimistic and pessimistic modes of TOP, respectively. TOP's authors selected these penalization levels from a vast choice of settings evaluated on the OpenAI Gym tasks. We name the resulting algorithm \textit{GPL-SAC-TOP}.

\textbf{Results.} In Figure~\ref{figure:e2e_opt} we show the performance curves for the evaluated alternative optimization procedures as compared to the original dual TD-learning in GPL-SAC:

GPL-SAC-E2E optimizes the uncertainty regularizer too aggressively, leading to instabilities and suboptimal performance across all tasks. We motivate these results based on two related inconsistencies of end-to-end bias minimization from Eqn.~\ref{eqn:e2e_bias_min}. Firstly, this alternative optimization does not consider the effects of the target action-value bias on the changes in the action distribution used for bootstrapping. In particular, modifying the uncertainty regularizer will affect the objective of the policy's optimization from Eqn.~\ref{pi_pg_obj} and the corresponding distribution of on-policy actions. Secondly, even considering a fixed policy, this end-to-end strategy assumes that $\beta$ linearly affects the bias. However, this is not the case due to the non-linear compounding effects from the bootstrapping in the TD-recursions. Hence, as empirically confirmed by the superior performance, framing bias minimization as a dual problem is a more general and appropriate formulation.

On the other hand, GPL-SAC-TOP matches the performance of dual TD-learning on the \textit{Humanoid} task but severely underperforms on the other two tasks. For \textit{Humanoid}, TOP's optimization strategy quickly converges to sampling $\beta=\sqrt{2}$ consistently, as $\beta=0$ causes considerable instabilities. As observed in Appendix~\ref{ext:unc_reg}, this task benefits from highly penalized targets for stabilization, making this early convergence to a high $\beta$ effective. However, both \textit{Walker} and \textit{Ant} tasks benefit from moderate levels of pessimism, which the bandit optimization strategy fails to recover in the examined experience regime. This inefficiency stems from the fact that immediate episodic return improvement does not appear to correlate with overall learning efficiency for these tasks. Moreover, the slow update frequency and noisy nature of the controller feedback make GPL-SAC-TOP unable to dynamically address arising biases. These results further show the effectiveness of dual TD-learning to directly minimize a source of learning instability and enable consistently efficient learning.

\textbf{Extended comparison with TOP and additional considerations}. We would like to note that the original TOP algorithm does not claim to estimate and counteract actual overestimation bias, but rather to switch between optimistic and pessimistic penalties (for better exploration or stability) to maximize past performance. Furthermore, while dual TD-learning can estimate the current bias \textit{on-the-fly}, after each optimization step, TOP relies on the full episodic returns to update its bandit-based controller, which can only provide a single feedback signal after completing a whole episode. As shown in its original paper, the slow nature of these updates appear to greatly constraint the number of possible pessimism values that can be optimized.

We provide a further comparison with TOP on a representative subset of three DeepMind Control environments by modifying our GPL-DrQ algorithms to incorporate TOP's optimization procedure rather than dual TD-learning. As shown in Table~\ref{app:tab:top_dmc} we see that dual TD-learning visibly outperforms TOP, with especially significant gains on the more complex Quadruped run environment. In line with previous results, this further shows that dual TD-learning attains better stability than simple bandit-based optimization.

\subsection{Pessimism Annealing}
\label{appsub:pess_ann_gym}

\begin{figure*}[t]
  \centering
  \includegraphics[width=0.8\linewidth]{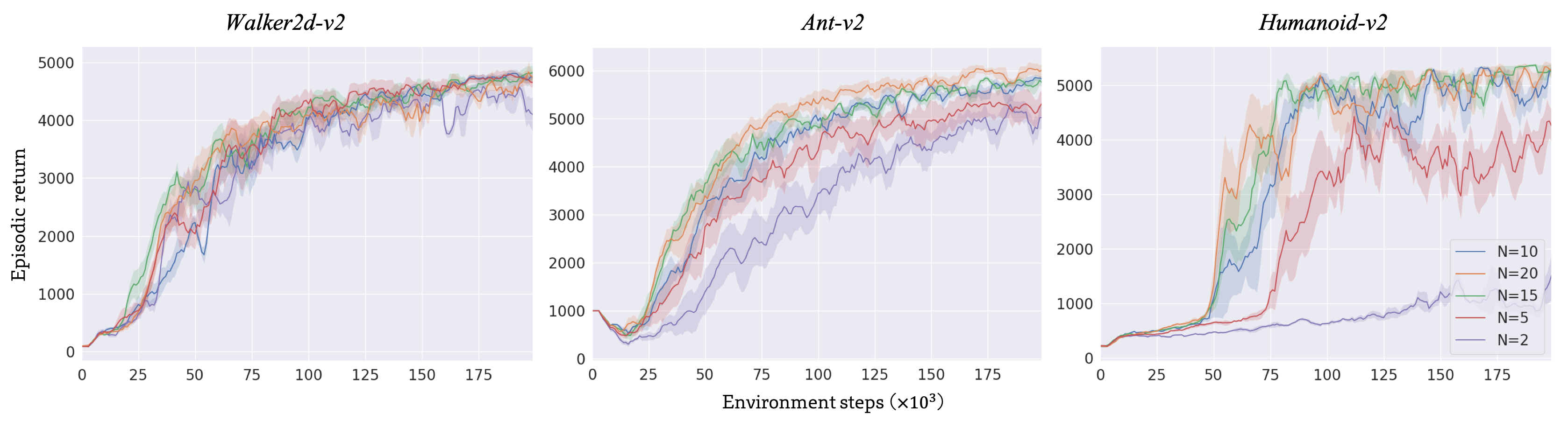}
  \caption{Performance curves showing the effects of varying the ensemble size of the critic in GPL-SAC.} 
  \label{figure:ens_size}
\end{figure*}
\begin{figure*}[t]
  \centering
  \includegraphics[width=0.8\linewidth]{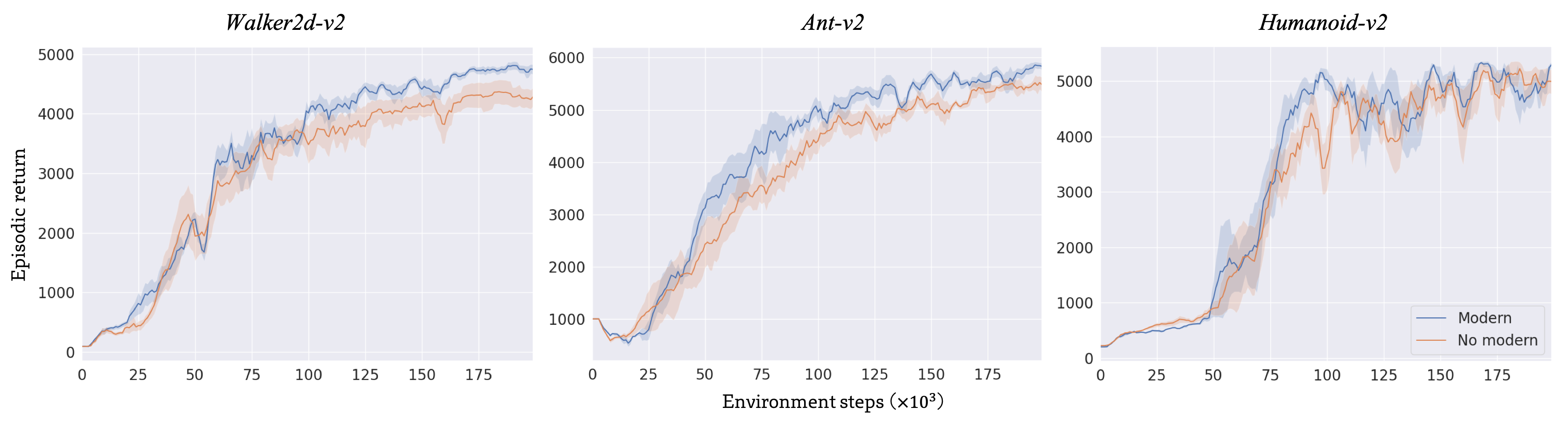}
  \caption{Performance curves showing the effects of substituting the modern architecture in GPL-SAC's action-value models with a standard fully-connected network.} 
  \label{figure:mod_arch}
\end{figure*}

We experiment with applying pessimism annealing to GPL-SAC for the OpenAI Gym tasks. In the same spirit as our previously described application for pixel observations tasks, we linearly decay $\lambda_{opt}$ from 0.5 to 0.0 in the first 50000 environment steps. We leave the rest of the hyper-parameters unaltered. We name the resulting algorithm \textit{GPL-SAC-Anneal}.

\textbf{Results.} In Figure~\ref{figure:pess_ann} we provide the performance curves comparing GPL-SAC with GPL-SAC-Anneal. The shifted uncertainty regularizer marginally improves performance in only two out of three environments. These results indicate that the undirected Gaussian exploration from SAC combined with the unbiased targets from GPL are already sufficient to ensure effective exploration in the OpenAI Gym tasks. In contrast, pessimism annealing appears to have a more significant effect in our results for the DeepMind Control Suite from Section~\ref{sec:experiments} of the main text, highlighting the harder exploration challenge introduced by these pixel-based tasks.

Furthermore, we also evaluate the effects of doubling or halving the initial optimistic shift value of $\lambda_{opt}$ from 0.5 to either 1 or 0.25 on a representative subset of three environments from the DeepMind Control suite. We keep the same linear annealing scheme of 500000 steps, following the standard deviation annealing in DrQv2, and the remaining hyper-parameters unaltered. We compare with the original GPL-DrQ-Anneal with  $\lambda_{opt}=0.5$ and GPL-DrQ (i.e., $\lambda_{opt}=0$).

\textbf{Results.} In Table~\ref{app:tab:optsh} we provide the average performance for the different initial values of $\lambda_{opt}$. Doubling $\lambda_{opt}$ appears to yield a slight boost in final performance for some of the harder exploration environments, such as Quadruped run. However, this comes at a slight performance cost in the remaining environments and slower overall learning, as shown by the inferior performance midway through training. These downsides are likely a side effect of optimizing the policy with an increasingly biased objective. In particular, while this practice might help exploring increasingly informative states, it naturally comes at the immediate cost of expected performance. Instead, halving the value of $\lambda_{opt}$ to 0.25 appears to yield some small improvements on Hopper, balanced by small performance decline on Quadruped. Overall performance is very similar to the original GPL-DrQ-Anneal with $\lambda_{opt}=0.5$.

\subsection{Critic Model Parameterization}
\label{appsub:ens}

We evaluate the performance of GPL-SAC with different parameterization of the critic's model. First, we examine the effect of the critic's ensemble size, comparing instances of GPL-SAC with $N=2$, $N=5$, $N=10$, $N=15$, and  $N=20$ action-value models. Second, we examine the effects of the modern action-value network architecture as compared to the seminal 3-layer fully-connected neural network employed in SAC. These parameters directly influence the ability to capture the critic's epistemic uncertainty, thus, affecting both the accuracy of the action-value predictions and the variance in the Wasserstein distance estimation for the uncertainty regularizer penalty.

\textbf{Results.} In Figure~\ref{figure:ens_size} we provide the performance curves for the different ensemble sizes. Predictably, both learning efficiency and final performance monotonically improve with the ensemble size. Yet, these improvements appear to saturate at different values of $N$ based on the complexity of the underlying environment, showing how harder tasks increasingly rely on accurately representing epistemic uncertainty. In particular, for \textit{Walker2d} training with only $N=2$ action-value models yields very similar results to training with $N=20$ action-value models. However, the performance differences are increasingly noticeable for the other tasks that involve significantly larger state and action spaces. For instance, for \textit{Humanoid} the agent is not able to recover meaningful behavior with $N=2$ and performance saturates only around training with $N=10$. In Figure~\ref{figure:mod_arch}, we provide the performance curves comparing GPL-SAC with and without our modern architecture. There appear to be consistent marginal improvements, with no significant performance differences in any of the considered environments. We opted to keep this alternative architecture in our final implementation of GPL-SAC since it did not introduce significant auxiliary costs in computational resources.



\subsection{Update-to-Data Ratio}
\label{appsub:utd}
\begin{figure*}
  \centering
  \includegraphics[width=0.8\linewidth]{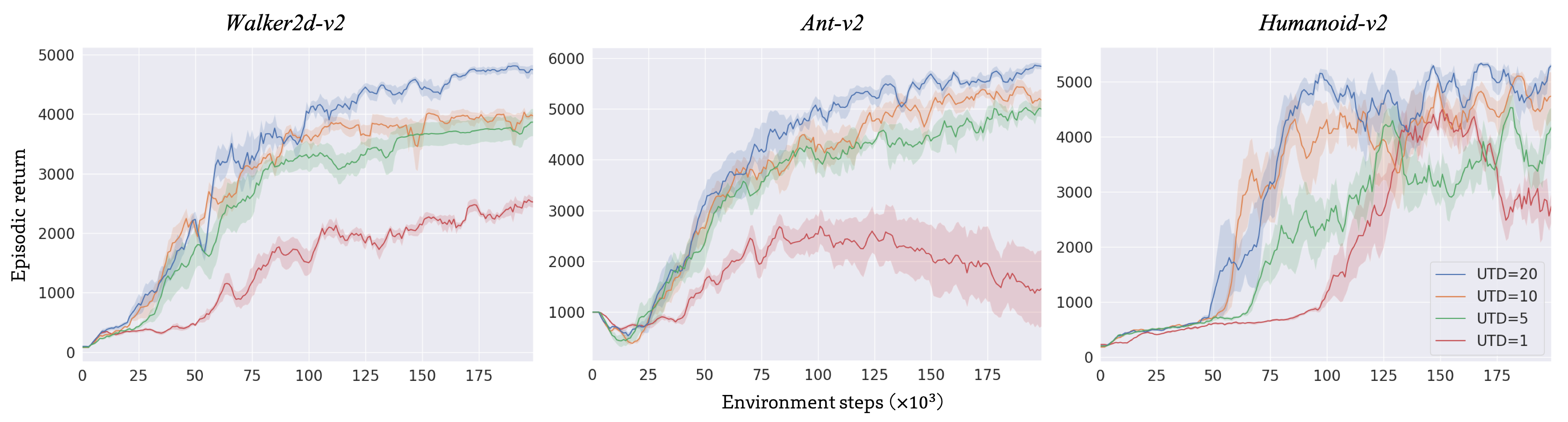}
  \caption{Performance curves showing the effects of lowering the critic's update-to-data ratio in GPL-SAC.} 
  \label{figure:utd}
\end{figure*}

Following the examined prior state-of-the-art algorithms, GPL-SAC uses a critic UTD ratio of $20$ across the different OpenAI Gym tasks. This aggressive optimization frequency ensures that the critic encapsulates most information from experience collected at any given point in the RL process. However, since the UTD ratio affects training time almost linearly, we examine its impact on performance. In particular, we compare instances of GPL-SAC with UTD ratios of $1$, $5$, $10$, and $20$.

\textbf{Results.} In Figure~\ref{figure:utd} we provide the performance curves for the different examined UTD ratios. Larger UTD ratios yield clear sample efficiency improvements for all tasks. Moreover, using a UTD ratio of $1$ in the \textit{Ant} and \textit{Humanoid} environments appears to cause some learning instabilities towards the end of the examined experience regime. These instabilities are likely from the inability of the learning process to quickly incorporate new information from recently collected data into the critic, given the growing size of the replay buffer. Consequently, a slow learning process might produce inaccurate return predictions for on-policy behavior, leading to a noisy policy gradient signal and, therefore, a degradation of the policy itself.


\subsection{Distributional Critic Parameterizations}
\label{appsub:dist_cri}
\begin{figure*}
  \centering
  \includegraphics[width=0.8\linewidth]{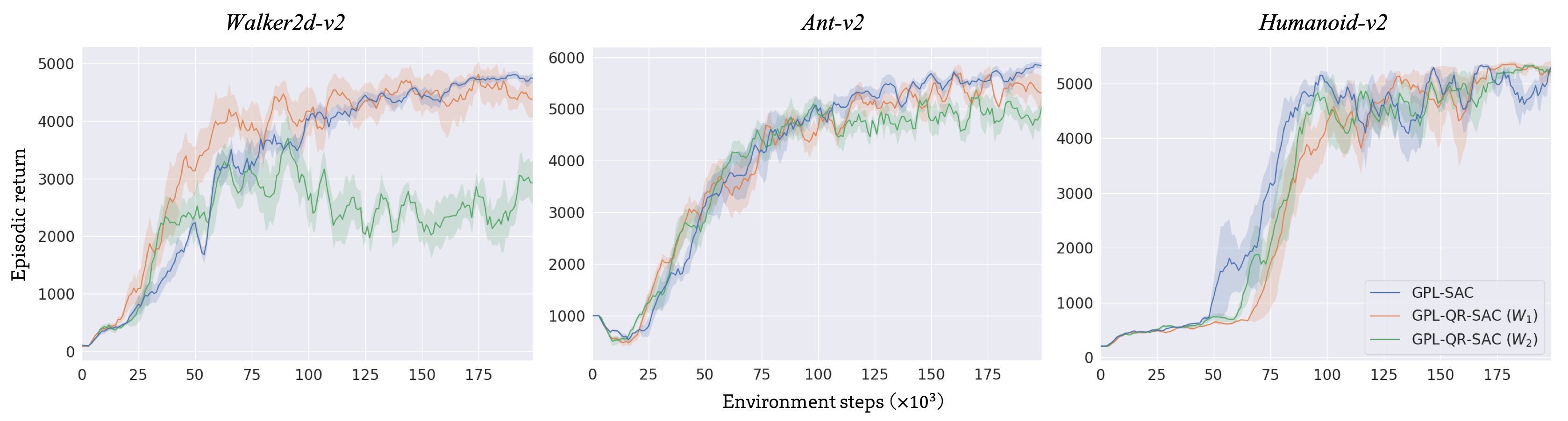}
  \caption{Performance curves showing the effects of employing a more expressive distributional critic based on quantile regression in GPL-SAC.} 
  \label{figure:quant}
\end{figure*}

The uncertainty regularizer penalty is a function of the expected Wasserstein distance between the return distributions predicted by the critic. Hence, it has a natural generalization to more expressive distributional critics beyond ensembles of action-value functions. Therefore, we evaluate the effectiveness of extending GPL-SAC by parameterizing the critic using an ensemble of $N$ distributional models $\{Z^\pi_{\phi_i}\}_{i=1}^N$. Following \citet{distributional-qrdqn}, we let each model output $M$ quantile locations values parameterizing a quantile distribution. We consider both 1-Wasserstein and 2-Wasserstein distances for the uncertainty regularizer (which for Dirac delta functions are equivalent). For quantile distributions, these distance measures are calculated as follows:
\begin{equation}\small
    \begin{split}
        W_1(Z^\pi_{\phi_i}, Z^\pi_{\phi_j}) = \sum_{k=1}^M \frac{1}{M}|Z^\pi_{\phi_i}(s, a)_k - Z^\pi_{\phi_j}(s, a)_k|,\\
        W_2(Z^\pi_{\phi_i}, Z^\pi_{\phi_j}) = \sqrt{\sum_{k=1}^M \frac{1}{M}\left(Z^\pi_{\phi_i}(s, a)_k - Z^\pi_{\phi_j}(s, a)_k\right)^2}.
    \end{split}
\end{equation}
We propose applying the uncertainty regularizer to each quantile location to compute the target return distribution, $\hat{Z}^\pi_{\phi}(s, a)\in \R^M$, as:
\begin{equation}\small
    \begin{split}
        \hat{Z}^\pi_{\phi}(s, a)_j = &Z_{\phi'}^\pi(s, a)_j - p_\beta(s', a', \phi,\theta), \\
        \textit{for all}& \quad j\in \{1, 2 \dots M\}.
    \end{split}
\end{equation}
We optimize the critic's models with a distributional version of TD-learning \citep{distributional-q-belle} making use of Huber quantile regression \citep{quantreg}. We still optimize the policy to maximize the expected returns by averaging over the quantile locations of the target return distribution. Moreover, we still perform dual TD-learning with minimal overheads by averaging all quantile errors computed for quantile regression during the distributional TD-learning updates. In our implementation, we approximate the return distribution with $M=10$ quantile locations and set the Huber quantile regression coefficient to $\kappa=1$. We keep all other hyper-parameters and model architectures consistent with GPL-SAC. We name the resulting algorithms \textit{GPL-QR-SAC} ($W_1$/$W_2$).

\textbf{Results.} In Figure~\ref{figure:quant} we provide the performance curves comparing both versions of GPL-QR-SAC with the original GPL-SAC algorithm. GPL-QR-SAC ($W_1$) appears to outperform GPL-QR-SAC ($W_2$), indicating that the 1-Wasserstein distance is a more effective and stable measure of epistemic uncertainty for $p_\beta$. However, in the examined tasks, GPL-QR-SAC ($W_1$) performs similarly to GPL-SAC, meaning there are no major benefits in parameterizing the critic with more expressive approximations to the return distribution. Yet, these results could also indicate that the returns for the considered OpenAI Gym environments have low stochasticity or that the action-value models already encapsulate all learnable information in the considered training regime. We leave further experimentation and analysis of using GPL with distributional critics for future work.

\subsection{Computational Scaling}
\label{appsub:comp_scaling}
\label{ext:comp_scaling}

\begin{table*}[t]
\caption{Average training times for the tested algorithms and ablations}
\label{tab:train_time}
\begin{center}
\begin{tabular}{@{}lc@{}}
\toprule
\textbf{Proprioceptive observations tasks} & Training time (seconds/1000 env. steps)  \\ \midrule
GPL-SAC ($N=10$, $UTD=20$)                   & $76.1$                       \\ \midrule
No pessimism learning                      & $74.2$                       \\
No modern action-value model               & $51.6$                       \\ \midrule
GPL-SAC-Anneal                              & $77.0$                       \\ \midrule
GPL-QR-SAC ($W_1$)                         & $86.6$                       \\
GPL-QR-SAC ($W_2$)                         & $88.1$                       \\ \midrule
$N=20$                                     & $85.0$                       \\
$N=15$                                     & $81.2$                       \\
$N=5$                                      & $62.5$                       \\
$N=2$                                      & $38.8$                       \\ \midrule
$UTD=10$                                   & $41.4$                       \\
$UTD=5$                                    & $20.1$                       \\
$UTD=1$                                    & $6.3$                        \\ \midrule
REDQ (Original implementation)             & $183.8$                      \\ \midrule
\textbf{Pixel observations tasks}          & Training time (seconds/10000 env. steps) \\ \midrule
GPL-DrQ-Anneal                              & $112.2$                      \\
GPL-DrQ                                    & $111.3$                      \\
DrQv2                                      & $111.2$                      \\ \bottomrule
\end{tabular}
\vspace{-10pt}
\end{center}
\end{table*}

We analyze the computational scaling of our implementations of GPL-based algorithms. Particularly, we record the average training time of executing the different considered algorithms for either 1000 environment steps for OpenAI Gym tasks or 10000 environment steps for DeepMind Control Suite tasks. We run each algorithm on an \textit{NVIDIA RTX 3090} GPU and an \textit{AMD Ryzen Threadripper 3970x} CPU. As described in Section~\ref{sec:experiments} of the main text, our implementation groups the parameters of the different models in the critic's ensemble into a single network to achieve better scaling for inference and backpropagation with distributed hardware. 

\textbf{Training times.} We report all average training times in Table~\ref{tab:train_time}. Comparing the training times with and without dual TD-learning and pessimism annealing clearly shows that both procedures introduce trivial computational overheads. Since the critic's updates represent the bulk of the computation in off-policy RL, modifying the associated hyper-parameters and parameterizations has relevant effects on training times. In particular, using a classical action-value model architecture speeds up training time by approximately $30\%$, while using a distributional critic slows down training by approximately $20\%$. Thanks to our implementation and the efficiency of large tensor operations, increasing the number of action-value models affects training times sub-linearly. For instance, doubling the critic's ensemble size to 20 results in less than $12\%$ additional training overhead. On the other hand, increasing the UTD ratio increases the total training time almost linearly. As compared to the original REDQ implementation from \citet{redq}, GPL-SAC trains in less than half the time while having similar algorithmic complexity.

\textbf{Considerations.} The results from this Section can provide intuitions on selecting components and hyper-parameters when applying our implementation of Generalized Pessimism Learning for different problems. For instance, increasing ensemble size beyond $N=10$ appears to affect training times only marginally. Thus, indicating that a viable criterion for selecting $N$ in practical scenarios could be based on how many models fit into GPU memory. Multiple directions can be explored to further improve the time-efficiency and scalability of GPL. Since our uncertainty regularizer is compatible with variational representations of the critic's parameters posterior, Bayesian parameterizations could be a viable option to capture epistemic uncertainty more efficiently than model ensembles. Moreover, recent work by \citet{halfprecision} shows that it is possible to train modern reinforcement learning algorithms in half precision, with non-trivial computation and memory benefits.

\section*{Ethical Statement}

\label{app:ESocImp}

Our work proposed a new method to improve the stability and efficiency of general off-policy reinforcement learning algorithms. Current applications of our method are confined to popular simulation benchmarks, yet, we recognize our advancements may contribute to future effective deployments of autonomous agents to tackle real-world problems. While, if unregulated, furthering automation could accentuate inequalities and have a tangible environmental impact, we believe potential societal benefits currently outweigh these concerns for the foreseeable future.

\end{document}